\documentclass[USenglish,oneside,twocolumn]{article}
\pdfoutput=1

\usepackage[big]{dgruyter_NEW}
\PassOptionsToPackage{sort&compress,numbers}{natbib}


\journalyear{2022}
\journalissue{1}

\author[1]{Bogdan Kulynych}
\author[2]{Mohammad Yaghini}
\author[3]{Giovanni Cherubin}
\author[4]{Michael Veale}
\author[1]{Carmela Troncoso}
\affil[1]{EPFL}
\affil[2]{University of Toronto, Vector Institute}
\affil[3]{Alan Turing Institute}
\affil[4]{University College London}

\usepackage[T1]{fontenc}    
\usepackage{booktabs}       
\usepackage{amsfonts}       
\usepackage{nicefrac}       
\usepackage{microtype}      
\usepackage{dsfont}
\usepackage{graphicx}
\usepackage{xcolor}
\usepackage{xspace}
\usepackage{subcaption}
\usepackage{float}
\usepackage{amsmath}
\usepackage{amssymb}
\usepackage{amsthm}
\usepackage{enumitem}
\usepackage{hyperref}
\usepackage[capitalise]{cleveref}
\usepackage{wrapfig}
\usepackage{libertine}
\usepackage[advantage,probability]{cryptocode}

\usepackage{microtype}
\usepackage{graphicx}
\usepackage{booktabs} 
\usepackage{tabularx}
\usepackage{framed}
\usepackage{algorithm}

\usepackage{pgfplots}
\pgfplotsset{compat=1.15}
\usepackage{mathrsfs}
\usetikzlibrary{arrows}

\newcommand{\newterm}[1]{{\it #1}}

\def\eqref#1{equation~\ref{#1}}

\def\1{\bm{1}}
\newcommand{\train}{\mathcal{D}}

\def\rl{{\textnormal{l}}}

\DeclareMathAlphabet{\mathsfit}{\encodingdefault}{\sfdefault}{m}{sl}
\SetMathAlphabet{\mathsfit}{bold}{\encodingdefault}{\sfdefault}{bx}{n}

\def\sW{{\mathbb{W}}}

\def\sZ{{\mathbb{Z}}}

\newcommand{\E}{\mathbb{E}}

\DeclareMathOperator*{\argmax}{arg\,max}

\newtheorem{corollary}{Corollary}

\newtheorem{proposition}{Proposition}

\theoremstyle{definition}
\newtheorem{definition}{Definition}
\theoremstyle{definition}

\newcommand{\adult}{\textsc{adult}\xspace}
\newcommand{\texas}{\textsc{texas-50K}\xspace}

\renewcommand{\newterm}{\textit}
\createprocedureblock {procb}{ center, boxed}{}{}{ linenumbering }

\newif\ifdraft
\drafttrue

\ifdraft
\newcommand{\bknote}[1]{{\color{cyan} \textbf{BK}: #1}}
\newcommand{\ctnote}[1]{{\color{magenta} \textbf{CT}: #1}}
\newcommand{\mynote}[1]{{\color{red} \textbf{MY}: #1}}
\newcommand{\notegio}[1]{{\color{orange} \textbf{GC}: #1}}
\else
\newcommand{\bknote}[1]{}
\newcommand{\updateme}[1]{}
\newcommand{\ctnote}[1]{}
\newcommand{\mynote}[1]{}
\newcommand{\notegio}[1]{}
\fi

\definecolor{darkgreen}{rgb}{0.0, 0.5, 0.0}
\definecolor{darkred}{rgb}{0.55, 0.0, 0.0}

\usepackage[normalem]{ulem}
\newcommand\soutpars[1]{\let\helpcmd\sout\parhelp#1\par\relax\relax}
\long\def\parhelp#1\par#2\relax{%
	\helpcmd{#1}\ifx\relax#2\else\par\parhelp#2\relax\fi%
}

\newcommand{\miagame}{\textsf{MIA}\xspace}
\renewcommand{\E}{\mathop{{}\mathbb{E}}}

\newcommand*\diff{\mathop{}\!\mathrm{d}}

\renewcommand{\train}{A}
\newcommand{\knowledge}[1][\obs]{\phi_{#1}}
\newcommand{\customfeatures}{\phi}
\newcommand{\define}{\triangleq}
\newcommand{\id}{\mathds{1}}
\newcommand{\loss}{\ell}
\newcommand{\inset}{S}
\newcommand{\outset}{\bar{\inset}}

\newcommand{\subgroup}{G}

\newcommand{\sample}{\xleftarrow{\$}}
\newcommand{\zset}{\sZ}

\newcommand{\att}{\mathtt{Att}}
\newcommand{\measure}[1][\yhat]{\mu^{#1}}
\newcommand{\propfunc}{\pi}

\newcommand{\vuln}[1][\obs]{\ensuremath{V}(\adv^*_{#1})\xspace}
\newcommand{\vulnz}[1][\obs]{\ensuremath{V}_{\zval}(\adv^*_{#1})\xspace}
\newcommand{\vulnzprim}[1][\obs]{\ensuremath{V}^*_{\zval'}(\adv^*_{#1})\xspace}
\newcommand{\deltavuln}[1][\obs]{\Delta \ensuremath{V}_{\zval,\zval'}(\adv^*_{#1})\xspace}

\newcommand{\vulnadv}[1][]{\ensuremath{V} #1\xspace}
\newcommand{\vulnadvz}[1][]{\ensuremath{V}_\zval #1\xspace}

\newcommand{\estvulnadv}[1][]{\ensuremath{\smash{\hat V}} #1\xspace}

\newcommand{\gap}{\mathsf{gap}\xspace}
\newcommand{\overfit}[2]{R({#1}, {#2})}
\newcommand{\overfitz}[3]{R_{#3}({#1}, {#2})}
\newcommand{\TV}{d_\mathrm{TV}}
\newcommand{\MD}{d_\mathrm{MD}}
\newcommand{\tvoverfit}[1][\knowledge]{\overfit{#1}{\TV}}
\newcommand{\tvoverfitz}[2][\knowledge]{\overfitz{#1}{\TV}{#2}}
\newcommand{\stoverfit}{\overfit{\loss}{\MD}}

\newcommand{\x}{X}
\newcommand{\y}{Y}
\newcommand{\z}{Z}

\newcommand{\yhat}{\smash{\hat Y}}
\newcommand{\obs}{W}
\newcommand{\obsval}{w}
\newcommand{\obspace}{\sW}
\newcommand{\ismember}{M}

\newcommand{\zval}{z}

\newcommand{\xval}{x}
\newcommand{\yval}{y}
\newcommand{\yhatval}{\smash{\hat y}}

\newcommand{\ismemberval}{m}

\newcommand{\dtrz}{{\substack{\inset \sim \datagen^n \\ \xval \sim (\inset \mid \zval)}}}
\newcommand{\dtrzprim}{{\substack{\inset \sim \datagen^n \\ \xval \sim (\inset \mid \zval')}}}
\newcommand{\dtez}{{\substack{\inset \sim \datagen^n \\ \xval \sim (\datagen \mid \zval)}}}
\newcommand{\dtezprim}{{\substack{\inset \sim \datagen^n \\ \xval \sim (\datagen \mid \zval')}}}

\newcommand{\clf}[1][\inset]{A_{#1}}
\newcommand{\datagen}{\mathcal{D}}
\newcommand{\pop}{\Omega}

\newcommand{\adv}{\mathcal{A}}
\newcommand{\advbayes}{\adv^*}
\newcommand{\attbayes}{\att^*}

\renewcommand{\paragraph}[1]{\medskip \noindent\textbf{#1.} }

\def\equationautorefname~#1\null{Equation~(#1)\null}

\title{\huge Disparate Vulnerability to Membership Inference Attacks \vspace{0.5em}}
\runningtitle{Disparate Vulnerability to Membership Inference Attacks}

\date{}

\begin{document}

\begin{abstract}
    {
    A membership inference attack (MIA) against a machine-learning model enables an attacker to determine whether a given data record was part of the model's training data or not. In this paper, we provide an in-depth study of the phenomenon of \emph{disparate
    vulnerability} against MIAs: unequal success rate of MIAs against different population subgroups.
    We first establish necessary and sufficient conditions for MIAs to be prevented, both on average and for population subgroups, using a notion of distributional generalization. Second, we derive connections of disparate vulnerability to algorithmic fairness and to differential privacy. 
    We show that fairness can only prevent disparate vulnerability against limited classes of adversaries. Differential privacy bounds disparate vulnerability but can significantly reduce the accuracy of the model.
    We show that estimating disparate vulnerability to MIAs by na\"ively applying existing attacks can lead to overestimation. 
    We then establish which attacks are suitable for estimating disparate vulnerability, and provide a statistical framework for doing so reliably.
    We conduct experiments on synthetic and real-world data finding statistically significant evidence of disparate vulnerability in realistic settings.
    }
\end{abstract}

\maketitle


\section{Introduction}
Membership Inference Attacks (MIAs), in which an adversary aims to determine whether an example is part of the training set, are one of the main privacy attacks against machine-learning (ML) models.
Since they were first described~\cite{ShokriSSS17}, many works have studied the potential of these attacks under diverse circumstances~\cite{ObermeyerE16,LumIsaac16,jayaraman2020revisiting,KhandaniKL10,NasrSH19,LongBWBWTGC18}; and the causes and limits of these attacks~\cite{farokhi2020modelling,LeinoF20,YeomGFJ18}.
In both empirical and theoretical approaches researchers focus on the \textit{average} MIA success across the records.
However, there is empirical evidence that the vulnerability to MIAs is not always evenly distributed: it can differ across target classes~\cite{ShokriSSS17}, it can be more effective against some individuals~\cite{LongBWBWTGC18}, and it can vary across subgroups~\cite{ChangS21}.
These results imply that average-based studies can overestimate the privacy for some individuals~\cite{EkstrandJM18}. 

In this paper, we provide the first theoretical analysis of the \textit{disparate vulnerability} to MIA across populations subgroups. 
Our contributions are the following:
\begin{itemize}
\item[\checkmark] We introduce a novel characterization of the vulnerability to MIAs, which provides a \emph{necessary and sufficient} condition for these attacks to succeed: lack of \emph{distributional generalization}. Vulnerability to MIA arises when the \emph{distribution} of a model's property (e.g., loss, or outputs) is different for samples in and out of the training dataset. This result complements previous studies that demonstrated the lack of standard generalization (i.e., overfitting) to be a sufficient but not necessary condition for vulnerability to MIAs~\cite{LongBWBWTGC18,YeomGFJ18}.

\item[\checkmark] We introduce the first formal analysis of disparate vulnerability and extend our results on necessary and sufficient conditions for preventing MIAs to subgroups.

\item[\checkmark] We show that estimating the magnitude of the disparate vulnerability is non-trivial when subgroups are small. We provide a statistical framework and methods to estimate disparate vulnerability and its significance. We show that not all vulnerability estimation mechanisms used in prior work are adequate for subgroups. We discuss the implications of these difficulties for regulation compliance.

\item[\checkmark] We prove that satisfying algorithmic-fairness constraints can decrease disparate vulnerability to limited classes of attackers.
We also show that training with differential privacy bounds the magnitude of the disparate vulnerability.

\item[\checkmark] We empirically evaluate disparate vulnerability both on synthetic and on real-world datasets, 
demonstrating that disparate vulnerability exists in realistic models, with high statistical significance.

\item[\checkmark] We discuss the importance of disagreggating privacy measurements when evaluating the legal implications of privacy attacks. In particular, the importance of studying the consequences of privacy attacks for subgroups when analyzing the privacy risks of a deployment, as opposed to studying individual privacy risks~\cite{LongBWBWTGC18} that can be dismissed as residual and acceptable.

\end{itemize}

\section{Related work}

\paragraph{Theory studies on MIA}
Yeom et al. studied the relation of MIAs to overfitting~\cite{YeomGFJ18}; in their work, they formalize MIA as an indistinguishability game, which we adapt to construct our theoretical framework.
Farokhi et al. analyzed the dependence of MIA's success on the amount of information the model memorizes~\cite{farokhi2020modelling}, and Jayaraman et al. investigated their dependence on the
prior probability that the example given to the adversary is a member or non-member of the training set~\cite{jayaraman2020revisiting}. \citet{YeomGFJ18}, and \citet{CherubinCP19} showed that MIAs success is bounded by DP. \citet{HumphriesRTOGK20} showed these bounds only apply so long as the training data are i.i.d.-sampled. All these analyses, however, are only
meaningful for the \textit{average}-case MIA. A classifier thought to be secure according to these analyses may provide weaker protection to certain individuals or subpopulations.

Our framework complements these studies and generalizes the notion of MIA risk to \textit{subgroups} of the population, thus enabling the  study of vulnerability for subsets of the records' labels, individuals, and subpopulations.

\paragraph{Disparity and machine learning}
The work on disparity in machine learning is centered on understanding and mitigating disparate impact of algorithmic decisions on subpopulations \cite{chouldechova_fair_2016,barocas_big_2016,LiptonMC18}.
\citet{BagdasaryanPS19} and \citet{PujolMKHMM20} study disparity in accuracy under differential privacy (DP), and show that training with DP can increase disparate impact.

In this work, we develop a theory that supports the empirical evidence that disparate impact would also cause disparity in vulnerability to MIAs~\cite{ChangS21,LongBWBWTGC18,ShokriSSS17}.

\section{Membership Inference Attacks}
\label{sec:background}



Let $\pop$ be a \newterm{population} of examples, where each example represents an individual: $\xval \in \pop$.
We assume that the population is partitioned in disjoint \newterm{subgroups}. Each subgroup $\subgroup_\zval \subset \pop$ is formed by examples that share one or several attributes (e.g., race or gender
in the way they are commonly represented in data),
such that $\bigcup_{\zval = 1}^t \subgroup_\zval = \pop$. We consider a \newterm{data-generating distribution} $\datagen$ 
over $\pop$.

We indicate with $\train(\cdot)$ the training algorithm that produces a model
$\clf$ given training data $\inset \subset \pop$. 
The learning task for this model is to infer the
value of the \newterm{label} $\yval = \yval(\xval)$ associated with an individual $\xval$.
We assume that the model can be either a regressor ($\yval$ takes values
in a set with total order, e.g. $\mathbb{R}$) or a classifier ($\yval$ takes values in a finite set).


The goal of a membership inference attack (MIA) is to predict whether an example $\xval \in \pop$ is a \newterm{member}
or a \newterm{non-member} of the training set $\inset$.
We assume a threat model where a MIA adversary observes the target model's behavior that relates to $\xval$, and has information about the data distribution $\datagen$, training-data sampling, and the training algorithm.
We formalize MIAs using the indistinguishability game by \citet{YeomGFJ18}:

\begin{figure}[h!]
\centering
\procb[linenumbering,space=auto]{$\miagame(\adv, \train, n, \datagen)$}{%
    \inset \gets \datagen^n; \clf = \train(S) \\
    \ismemberval \sample \{0,1\}\\
    \pcif \ismemberval = 1 \pcthen\\
        \xval \sample \inset\\
    \pcelse\\
        \xval \gets \datagen\\
    \pcendif\\
    \hat \ismemberval \gets \adv(\xval, \clf, n, \datagen\big)\\
    \pcreturn \ismemberval = \hat \ismemberval}
\end{figure}

In this game, the challenger samples $\inset$ from the population, and trains a model $\clf$ using training algorithm $\clf[]$ (line 1).
The challenger then randomly draws a secret $m$ (line 2) whose value denotes $x$'s $m$embership in $S$: $m=1$ if the \newterm{challenge example} $x$ is sampled from the training set $\inset$ (line 4), and $m=0$ if it is sampled from the data distribution $\datagen$ (line 6). As \citet{YeomGFJ18}, we assume that the population is large
enough that the chance of sampling a member $x \in \inset$ from $\datagen$ is negligible. Given the challenge example $x$, the target model $\clf$ and its training algorithm $\train(\cdot)$, the sampling parameter $n$, and the distribution of the training data $\datagen$, the MIA adversary $\adv(\cdot)$ makes a guess $\hat m$ about the example's membership in $\inset$ (line 8). 
We use this formalization as it is the most common, although there are other ways to formalize MIAs~\cite{HumphriesRTOGK20}. 

The $\miagame$ game defines a joint probability distribution over training datasets $\inset$, membership ``coins'' $\ismemberval$, and challenge examples $\xval$. We denote by $\ismember$ the random variable taking the value of the membership coin (line 2), by $\x$ the challenge example, by $\y = \yval(\x)$ the label associated with the challenge example $\xval$, by $\z$ the subgroup of the population $\zval$ to which the $\xval$ belongs, and by $\yhat = \clf(\x)$ the output the model $\clf\textbf{}$ at $\xval$.


\subsection{Attack strategy}
As described in the \miagame game, the adversary's
knowledge is limited to $(\xval, \clf, n, \datagen)$,
and their goal is to guess the membership of $\xval$. For brevity, we use $\clf$ to indicate both the access to trained models $\clf$ and their training algorithm $\train(\cdot)$. 

We define a general strategy to perform a membership attack that encompasses
several instances of MIA, e.g., \cite{ShokriSSS17,YeomGFJ18,NasrSH19}.
This strategy consists of two phases.

First, the adversary prepares an attack algorithm $\att_{\train, n,\datagen}(\cdot)$ which
depends on the target training algorithm $\train(\cdot)$, and data-sampling parameters $n$ and $\datagen$, e.g., by training a shadow-model attack classifier~\cite{ShokriSSS17}. We drop the subscripts in $\att_{\train, n, \datagen}$ where the setting is clear from the context.

In the second phase, the adversary extracts \newterm{features}, $\obsval \leftarrow \knowledge[](\clf, \xval)$, describing the target model and the example, and applies the attack algorithm to the extracted features to obtain the membership guess, $\hat \ismemberval \leftarrow \att_{\train, n, \datagen}(\obsval)$.
Thus, the adversary's guess $\hat \ismemberval$ is obtained
by applying the attack algorithm to the extracted
features:
\[
    \adv(\xval, \clf, n, \datagen) \define \att_{\train, n, \datagen} \circ \knowledge[](\clf, \xval)
\]

This formalization is flexible: it captures both white-box and black-box adversarial models.
For example, the features could be the outputs of the model and the example's label $w=(\clf(\xval), \yval(\xval))$ \cite{ShokriSSS17},
the model's loss $\ell$ for the challenge example,
$w=\ell(\clf(\xval), \yval(\xval))$ \cite{YeomGFJ18}, or the model's gradients as in some white-box attacks~\cite{NasrSH19}, etc. 

We use random variable $\obs$ to indicate the
extracted features $\obsval$ across instances of
the $\miagame$ game. For example, if the attacker uses the model's output and the label as features~\cite{ShokriSSS17}, we denote them as $\obs = (\yhat, \y)$.
With a slight abuse of notation, we use
$\knowledge: (\clf, \xval) \mapsto \obsval$ to
indicate the procedure that extracts features $\obsval$ that are realizations of the $\obs$ random variable.
Furthermore, 
we denote by $\adv_\obs$ an adversary that uses features $\obs$.


We distinguish two kinds of adversaries depending on the features they use: \textit{regular} adversaries that do not use subgroup information ($\z \notin \obs$), and \textit{subgroup-aware} adversaries that do use this information ($\z \in \obs$). We assume that the latter adversary can obtain the subgroup $z$ from the examples $x$ themselves, encoded in an example
(e.g., gender, race). That is the case for our experiments on real-world data in \cref{sec:realworld}.
However, in practical scenarios, this knowledge could
be encoded in the label $\yval(\xval)$, or come from external sources. Prior work has mainly considered regular adversaries.

\subsection{Vulnerability}
We introduce the concept of \newterm{vulnerability} of an ML model to membership inference attacks (MIAs). Vulnerability measures the success of an adversary against the model. We also introduce worst-case (Bayes) vulnerability, i.e., vulnerability against an information-theoretically optimal adversary.

We measure the vulnerability to MIAs by the normalized advantage~\cite{YeomGFJ18} of the adversary $\adv$ over random guessing:
\begin{definition}
	\label{def:vulnerability}
	We define \newterm{vulnerability} to adversary $\adv$ as:
	\begin{equation}\label{eq:vulnerability}
	\begin{aligned}
	    \vulnadv[(\adv)] &\triangleq 2\Pr[\miagame(\adv, \train, n, \datagen) = 1] - 1
	    \end{aligned}
	\end{equation}
\end{definition}

The definition of vulnerability can be extended to subgroups:
\begin{definition}
	Let $\zval$ be a subgroup of the population. We define \newterm{subgroup vulnerability} to adversary $\adv$ as:
	\[
        \vulnadvz[(\adv)] \triangleq 2 \Pr[\miagame(\adv, \train, n, \datagen) = 1 \mid \z = \zval ] - 1.
    \]
\end{definition}
which captures the normalized advantage of a MIA adversary $\adv$ for challenge examples coming from a given subgroup $\zval$.

\paragraph{Optimal adversaries}
We base our analysis on information-theoretically optimal adversaries. Consider the worst-case vulnerability to any attack that uses features $\obs$:
\begin{equation}\label{eq:bayes-vuln}
    \begin{aligned}
        \max_{\att_\obs: \obspace \mapsto \{0, 1\}} \vulnadv[(\att_\obs\,\circ\,\knowledge)],
    \end{aligned}
\end{equation}
where $\obspace$ is the domain of $\obs$.

The maximum in Eq.~\ref{eq:bayes-vuln} is achieved by a \newterm{Bayes adversary} which uses the following
strategy for the attack~\cite{CherubinCP19, SablayrollesDSO19}:
\begin{equation}\label{eq:adv-definition}
\begin{aligned}
    \attbayes_\obs(\obsval) &\triangleq \argmax_{\ismemberval \in \{0, 1\}} \Pr[\ismember = \ismemberval \mid \obs = \obsval],
\end{aligned}
\end{equation}
We denote the Bayes adversary as $\advbayes_\obs \define \attbayes_\obs \circ \knowledge$, and drop the subscripts where no ambiguity arises.

\paragraph{Subgroup-aware Bayes adversary}
We assume the adversary knows the subgroup $\zval$ to which each example $\xval$ belongs.
We refer to this adversary as subgroup-aware.
As the vulnerability to the Bayes adversary grows if the adversary has more information about the examples, the worst-case vulnerability to a subgroup-aware (Bayes) adversary is equal or higher compared to a regular (Bayes) adversary:
\begin{proposition}
\label{prop:discriminating-vs-regular}
	$\vuln[\obs, \z] \geq \vuln[\obs] \,.$
\end{proposition}
We defer the proof to \cref{sec:proofs}.

In our experimental evaluations, we only consider subgroup-aware adversaries as they are guaranteed to attain higher advantage in the worst-case.
\section{Distributional Generalization and Vulnerability to MIAs}
\label{sec:why-mia}

An ML model is said to \textit{overfit}, or poorly \textit{generalize}, when its average loss on the training set differs from its loss on new samples from the  population. Previous work showed that, while overfitting is an important factor for evaluating MIA~\cite{ShokriSSS17}, it is not necessary for MIA vulnerability~\cite{YeomGFJ18, LongBWBWTGC18}. 

\cref{fig:average-overfitting} illustrates with an example why the absence of standard overfitting does not, in general, prevent MIAs. The figure shows a model's loss values on its training and test data. The standard, average-based definition of overfitting cannot distinguish between the two distributions; but an adversary potentially can, and the model can be vulnerable to MIAs.

\subsection{Distributional Generalization}
To establish the necessary and sufficient conditions for models to be vulnerable to MIAs, we introduce an extended notion of generalization that goes beyond comparing the average loss on train and test data. It covers the difference in the distributions of any given property of a model on the training data and outside. A \newterm{property} is any function that takes as input a model and an example: $\propfunc(\clf, \xval)$, and returns a numeric vector. A property function can be, for instance, a loss function, the gradient, or the prediction from the model.

We are interested in the distributions of properties on the examples $\xval$ coming from the training dataset and from outside of the training dataset. For any set $T$ from the range of $\propfunc$, we define the corresponding probability measures as:
\[
    \begin{aligned}
        \measure[\propfunc]_1(T) &\define \Pr_{\substack{\inset \sim \datagen^n \\ \xval \sim \inset}}[\propfunc(\clf, \xval) \in T] \\ 
        \measure[\propfunc]_0(T) &\define \Pr_{\substack{\inset \sim \datagen^n \\ \xval \sim \datagen}}[\propfunc(\clf, \xval) \in T]
    \end{aligned}
\]

\begin{definition}
For any property function $\propfunc(\clf, \xval)$, we define \newterm{the distributional-generalization gap} as follows:
\[
    \overfit{\propfunc}{d} \define d\big(\measure[\propfunc]_1,\ \measure[\propfunc]_0\big),
\]
where $d(\mu, \mu')$ is a measure of dissimilarity between probability distributions.
\end{definition}

This generic notion subsumes the standard notion of generalization. Standard generalization can be measured using the average-dataset generalization gap (see, e.g., in \citet {YeomGFJ18}), the difference between the expected loss on the training dataset and the expected loss on the distribution:
\[
    \begin{aligned}
        R &\define \E_{\substack{\inset \sim \datagen^n \\ \xval \sim \inset}}[\loss(\clf, \xval)] - \E_{\substack{\inset \sim \datagen^n \\ \xval \sim \datagen}}[\loss(\clf, \xval)],
    \end{aligned}
\]
where $\loss(\clf, \xval)$ is a loss function. We can recover this standard notion of a generalization gap as $\stoverfit$, using the loss function as the property function and the \newterm{mean discrepancy} function $\MD(\mu, \mu')$ as a measure of dissimilarity:
\[
    \MD(\mu, \mu') \define \int \omega \diff \mu(\omega) - \int \omega \diff \mu'(\omega),
\]

Whereas standard generalization quantifies how much the training algorithm tends to memorize the training dataset through the lens of its performance (loss), distributional generalization can do so (1) through the lens of other properties beyond losses, and (2) considering distributional information instead of only the difference between the means.

Evaluating distributional generalization enables us
to assess the generalization of an ML model
on the entire population, rather than on average.
In \cref{fig:average-overfitting} 
it is clear that the model's
actual loss across the entire population
is concentrated on a few individuals.
Distributional generalization enables us to capture
this discrepancy, whereas standard generalization does not.

\begin{figure}
    \centering
    \tikzset{every picture/.style={line width=0.75pt}} 

\begin{tikzpicture}[x=0.75pt,y=0.75pt,scale=0.65,yscale=-1,xscale=1]

\draw  [color={rgb, 255:red, 208; green, 2; blue, 27 }  ,draw opacity=1 ][fill={rgb, 255:red, 208; green, 2; blue, 27 }  ,fill opacity=0.36 ] (31.58,500.67) -- (150.5,500.67) -- (150.5,509) -- (31.58,509) -- cycle ;
\draw  [color={rgb, 255:red, 74; green, 144; blue, 226 }  ,draw opacity=1 ][fill={rgb, 255:red, 74; green, 144; blue, 226 }  ,fill opacity=0.44 ] (229.33,509.67) -- (229.33,450) -- (249.5,450) -- (249.5,509.67) -- cycle ;
\draw [line width=1.5]    (30.58,509.03) -- (176,509.03) ;
\draw [shift={(180,509.03)}, rotate = 180] [fill={rgb, 255:red, 0; green, 0; blue, 0 }  ][line width=0.08]  [draw opacity=0] (11.61,-5.58) -- (0,0) -- (11.61,5.58) -- cycle    ;
\draw [line width=1.5]    (30.58,404.92) -- (30.58,509.03) ;
\draw [shift={(30.58,400.92)}, rotate = 90] [fill={rgb, 255:red, 0; green, 0; blue, 0 }  ][line width=0.08]  [draw opacity=0] (11.61,-5.58) -- (0,0) -- (11.61,5.58) -- cycle    ;
\draw [line width=1.5]    (201.08,510.53) -- (346.5,510.53) ;
\draw [shift={(350.5,510.53)}, rotate = 180] [fill={rgb, 255:red, 0; green, 0; blue, 0 }  ][line width=0.08]  [draw opacity=0] (11.61,-5.58) -- (0,0) -- (11.61,5.58) -- cycle    ;
\draw [line width=1.5]    (201.08,406.42) -- (201.08,510.53) ;
\draw [shift={(201.08,402.42)}, rotate = 90] [fill={rgb, 255:red, 0; green, 0; blue, 0 }  ][line width=0.08]  [draw opacity=0] (11.61,-5.58) -- (0,0) -- (11.61,5.58) -- cycle    ;

\draw (96.41,525) node    {$\inset$};
\draw (71,402.41) node    {$\ell ( A_{s} ,\ x)$};
\draw (270.41,525) node    {$\outset$};
\draw (242,403.91) node    {$\ell ( A_{s} ,\ x)$};

\end{tikzpicture}
    \vspace{-1em}
    \caption{Loss values of a model $\clf$
    on train data $\inset$ (left) and test data $\outset$ (right).
    According to standard notion of generalization, this model does
    not overfit:
    average loss (area) on training and test data is identical.
    Some population individuals, however, are more
    penalized on the test data.
    This discrepancy is captured by \emph{distributional} generalization.
    }
    \label{fig:average-overfitting}
\end{figure}

Concurrently, \citet{NakkiranBansal20} have also proposed a similar notion of distributional generalization. Our proposal allows for more general distances between distributions, whereas \citeauthor{NakkiranBansal20}, when translated to our terms, define the gap using the mean discrepancy, which is not sufficient for our analysis.

\subsection{Relation between Worst-case Vulnerability and Distributional Generalization}

The ability of any classifier to successfully distinguish between observations of
two classes can be characterized by the total variation between the class-conditional
distributions of observations. By applying this well-known fact to the worst-case MIA attackers, we can characterize vulnerability in terms of distributional generalization:
\begin{proposition}
\label{thm:vuln-tv}
    The worst-case vulnerability to MIAs with adversary's features $\obs$ is 
    equal to the distributional-generalization gap under total-variation distance:
    \[
        \begin{aligned}
            \vuln[\obs] &= \tvoverfit,
        \end{aligned}
    \]
    where the total-variation distance is defined as:
    \[
        \TV(\mu, \mu') \define \sup_{T \subseteq \obspace} |\mu(T) - \mu'(T)|
    \]
\end{proposition}

According to \cref{thm:vuln-tv}, when the property function $\pi$ is the adversary's feature extraction mechanism $\knowledge$, the distributional-generalization gap is equal to the worst-case vulnerability to adversaries that use features  $\obs = \knowledge(\clf, \x)$. 

\begin{proof}
    Let us define the \newterm{Bayes error} $L^*$, the 0-1 classification error of the Bayes classifier. In the case of $\attbayes$:
    \[
        L^* \define \Pr[\attbayes(\obs) \neq \ismember]
    \]
    Recall that vulnerability is defined through the success probability of an adversary:
    \[\vulnadv[(\adv_\obs)] \define 2\Pr[\att(\obs) = \ismember] - 1\]
    Thus, for a Bayes adversary, $\vuln$ uses the complement of the Bayes error $L^*$:
    \[\vuln = 2(1 - \Pr[\attbayes(\obs) \neq \ismember]) - 1 = 1 - 2 L^*.\]
    It is well-known that the the Bayes error of the binary classifier under uniform prior is equal to:
    \[
        \begin{aligned}
            L^* & = \frac{1}{2} - \frac{1}{2} \, \TV\left( \Pr[\obs \mid \ismember = 1],\ \Pr[\obs \mid \ismember = 0] \right) \\
                & = \frac{1}{2} - \frac{1}{2} \, \TV\hspace{-0.25em}\left( \Pr_{\substack{\inset \sim \datagen^n \\ \xval \sim \inset}}[\knowledge(\clf, \xval)],\ 
                \Pr_{\substack{\inset \sim \datagen^n \\ \xval \sim \datagen}}[\knowledge(\clf, \xval)] \right) \\
                & = \frac{1}{2} - \frac{1}{2} \, \TV\left(\measure[\knowledge]_1,\ \measure[\knowledge]_0\right),
        \end{aligned}
    \]
    See, e.g., ~\citet[Chapter~3.9]{DevroyeGL13}. This implies the sought form.

\end{proof}
This form is a straightforward consequence of our Bayes-optimal approach to vulnerability and is an
application of a well-known result in statistical theory. It provides us with an intuitive
interpretation of the worst-case vulnerability to MIAs---as it is equal to the distributional-generalization
gap---thus with a guideline on how to prevent MIAs. The result holds for both white-box and black-box adversary models.

Let us visually illustrate distributional generalization and worst-case vulnerability. Consider adversarial features $\obs = \yhat$. For the continuous property function $\customfeatures_{\yhat}$, the distributional-generalization gap becomes:
\[
    \newcommand{\knowledgeyhat}{\knowledge[\yhat]}
    \begin{aligned}
        \tvoverfit[\knowledgeyhat] &= \TV\left( \measure[\customfeatures_{\yhat}]_1, \measure[\customfeatures_{\yhat}]_0 \right) \\
                          &= \frac{1}{2} \int \big|f_1(\yhatval) - f_0(\yhatval)\big| \diff \yhatval,
    \end{aligned}
\]
where $f_1$ and $f_0$ are probability density functions associated with measures $\mu_1$ and $\mu_0$, respectively. See \cref{fig:bayes-overfit-illustration} for a visualization. The worst-case vulnerability to adversaries using features $\obs = \yhat$ is the area between the densities of the ``in'' and ``out'' output distributions.

Note that the distance used in \cref{thm:vuln-tv} is \emph{average-dataset}. That is, when computing the features $\phi(\clf, \x)$, the model $\clf$ is a random variable over the randomness of  $\train(\cdot)$ and $\inset \sim \datagen^n$. To train models with minimal vulnerability to MIAs, \citet{LiLR21} used a similar yet different notion of distance, the distance between outputs of a \emph{fixed} model on its training dataset and a validation dataset. Although conceptually similar, such distance cannot be directly used to evaluate the worst-case vulnerability using \cref{thm:vuln-tv}.

\begin{figure}
    \centering
    \usetikzlibrary{patterns}


\tikzset{
pattern size/.store in=\mcSize,
pattern size = 5pt,
pattern thickness/.store in=\mcThickness,
pattern thickness = 0.3pt,
pattern radius/.store in=\mcRadius,
pattern radius = 1pt}
\makeatletter
\pgfutil@ifundefined{pgf@pattern@name@_d2bxs13xo}{
\pgfdeclarepatternformonly[\mcThickness,\mcSize]{_d2bxs13xo}
{\pgfqpoint{-\mcThickness}{-\mcThickness}}
{\pgfpoint{\mcSize}{\mcSize}}
{\pgfpoint{\mcSize}{\mcSize}}
{
\pgfsetcolor{\tikz@pattern@color}
\pgfsetlinewidth{\mcThickness}
\pgfpathmoveto{\pgfpointorigin}
\pgfpathlineto{\pgfpoint{0}{\mcSize}}
\pgfusepath{stroke}
}}
\makeatother


\tikzset{
pattern size/.store in=\mcSize,
pattern size = 5pt,
pattern thickness/.store in=\mcThickness,
pattern thickness = 0.3pt,
pattern radius/.store in=\mcRadius,
pattern radius = 1pt}
\makeatletter
\pgfutil@ifundefined{pgf@pattern@name@_n9047vln7}{
\pgfdeclarepatternformonly[\mcThickness,\mcSize]{_n9047vln7}
{\pgfqpoint{-\mcThickness}{-\mcThickness}}
{\pgfpoint{\mcSize}{\mcSize}}
{\pgfpoint{\mcSize}{\mcSize}}
{
\pgfsetcolor{\tikz@pattern@color}
\pgfsetlinewidth{\mcThickness}
\pgfpathmoveto{\pgfpointorigin}
\pgfpathlineto{\pgfpoint{0}{\mcSize}}
\pgfusepath{stroke}
}}
\makeatother
\tikzset{every picture/.style={line width=0.75pt}} 

\begin{tikzpicture}[x=0.75pt,y=0.75pt,yscale=-1,xscale=1,thick,scale=0.65, every node/.style={scale=0.65}]

\draw [color={rgb, 255:red, 74; green, 144; blue, 226 }  ,draw opacity=1 ][pattern=_d2bxs13xo,pattern size=6pt,pattern thickness=0.75pt,pattern radius=0pt, pattern color={rgb, 255:red, 155; green, 155; blue, 155}]   (166.14,249.16) .. controls (211.52,249.76) and (215.44,168.73) .. (239.23,168.73) .. controls (263.03,168.73) and (263.94,244.93) .. (325.26,249.64) ;
\draw [color={rgb, 255:red, 208; green, 2; blue, 27 }  ,draw opacity=1 ][pattern=_n9047vln7,pattern size=6pt,pattern thickness=0.75pt,pattern radius=0pt, pattern color={rgb, 255:red, 155; green, 155; blue, 155}]   (208.28,249.5) .. controls (253.66,250.1) and (270.76,179.71) .. (294.56,179.71) .. controls (318.35,179.71) and (326.45,244.93) .. (387.76,249.64) ;
\draw [line width=1.5]    (159.58,144.92) -- (159.58,249.03) ;
\draw [shift={(159.58,140.92)}, rotate = 90] [fill={rgb, 255:red, 0; green, 0; blue, 0 }  ][line width=0.08]  [draw opacity=0] (11.61,-5.58) -- (0,0) -- (11.61,5.58) -- cycle    ;
\draw [draw opacity=0][fill={rgb, 255:red, 255; green, 255; blue, 255 }  ,fill opacity=1 ]   (212.86,249.5) .. controls (223.69,248.42) and (230.76,246.37) .. (244.91,233.18) .. controls (259.07,218.5) and (266.14,205.52) .. (266.56,205.52) .. controls (266.56,205.52) and (276.97,224) .. (288.62,235.08) .. controls (296.53,239.6) and (292.37,241.29) .. (317.76,249.07) ;
\draw [color={rgb, 255:red, 74; green, 144; blue, 226 }  ,draw opacity=1 ][fill={rgb, 255:red, 74; green, 144; blue, 226 }  ,fill opacity=0.25 ]   (166.14,249.16) .. controls (211.52,249.76) and (215.44,168.73) .. (239.23,168.73) .. controls (250.16,168.73) and (256.26,184.79) .. (265.6,202.62) .. controls (276.6,223.63) and (292.09,247.09) .. (325.26,249.64) ;
\draw [color={rgb, 255:red, 208; green, 2; blue, 27 }  ,draw opacity=1 ][fill={rgb, 255:red, 208; green, 2; blue, 27 }  ,fill opacity=0.25 ]   (208.28,249.5) .. controls (253.66,250.1) and (270.76,179.71) .. (294.56,179.71) .. controls (318.35,179.71) and (326.45,244.93) .. (387.76,249.64) ;

\draw [line width=1.5]    (159.58,249.03) -- (401.41,249.03) ;
\draw [shift={(405.41,249.03)}, rotate = 180] [fill={rgb, 255:red, 0; green, 0; blue, 0 }  ][line width=0.08]  [draw opacity=0] (11.61,-5.58) -- (0,0) -- (11.61,5.58) -- cycle    ;

\draw (356.71,200.51) node    {$M=0$};
\draw (185.13,199.95) node    {$M=1$};
\draw (414.41,233.46) node    {$\hat{y}$};
\draw (204.6,152.07) node    {$f_{m}(\yhatval)$};

\end{tikzpicture}
    \vspace{-1em}
    \caption{Distributional-generalization gap for models' outputs $\yhatval$. 
    The curves represent the probability density functions of models' outputs
    on the training datasets ($\ismember = 1$) and outside ($\ismember = 0$). The striped area shows the
    distributional-generalization gap: total variation between distributions of model's outputs on training and outside. \cref{thm:vuln-tv} shows that the
    the size of the striped area exactly equals to the worst-case vulnerability to any adversary that uses model outputs $\yhatval$ as features for distinguishing members from non-members.}
    \label{fig:bayes-overfit-illustration}
\end{figure}

\paragraph{Standard overfitting and worst-case vulnerability}
The absence of overfitting in the standard sense does not necessarily preclude
MIAs~\cite{YeomGFJ18, LongBWBWTGC18}.
But, a straightforward implication of \cref{thm:vuln-tv} shows there is a case when the standard generalization gap does bound the
worst-case vulnerability:
\begin{corollary}
    Let $\loss(\clf, \xval) = \id[\clf(\xval) \neq \yval(\xval)]$ be the 0-1 loss, and the adversary's
    features be the loss values $\obs = \loss(\clf, \x)$. Then, the standard generalization gap equals worst-case vulnerability:
    \begin{equation}\label{eq:dist-eq-standard-overfit}
        \begin{aligned}
                  \vuln[\loss(\clf, \x)] &= |\stoverfit|
        \end{aligned}
    \end{equation}
\end{corollary}
\begin{proof}
    As loss is binary-valued, $\tvoverfit[\loss]$ simplifies to:
    \[
        \begin{aligned}
            \tvoverfit[\loss] &= |\Pr[\loss(\clf, \x) = 1 \mid \ismember = 1] \\
                         &\quad- \Pr[\loss(\clf, \x) = 1 \mid \ismember = 0] | \\
                         &= |\E[\loss(\clf, \x) \mid \ismember = 1] \\
                         &\quad- \E[\loss(\clf, \x) \mid \ismember = 0)]| \\
                         &= |\stoverfit|.
        \end{aligned}
    \]
\end{proof}
Therefore, if a MIA adversary only observes whether a queried example has a correct or incorrect
prediction by the target model, the upper bound on the success of any such attack has a direct
relationship to standard overfitting $\stoverfit$. Thus, for such an
adversarial model, no overfitting \emph{does} imply no vulnerability to MIAs.

\subsection{Disparate Vulnerability}
\label{sec:disp-theory}
In this section, we provide a theoretical analysis of vulnerability to MIAs disaggregated by subgroups.

We introduce a subgroup-specific version of distributional generalization, in which the distributions of the property $\propfunc$ are computed on examples that belong to a given subgroup. For any set $T$ from the range of $\propfunc$, we define subgroup-specific measures:
\[
    \begin{aligned}
        \measure[\propfunc]_{1, \zval}(T) &\define \Pr_{\substack{\inset \sim \datagen^n \\ \xval \sim (\inset \mid \zval)}}[\propfunc(\clf, \xval) \in T] \\ 
        \measure[\propfunc]_{0, \zval}(T) &\define \Pr_{\substack{\inset \sim \datagen^n \\ \xval \sim (\datagen \mid \zval)}}[\propfunc(\clf, \xval) \in T],
    \end{aligned}
\]
where $\xval \sim (\cdot \mid \zval)$ denotes sampling conditioned on the subgroup $\zval$.

\begin{definition}
For any property function $\propfunc(\clf, \xval)$, we define \newterm{the subgroup-specific distributional-generalization gap}:
\[
    \overfitz{\propfunc}{d}{\zval} \define d\big(\measure[\propfunc]_{1, \zval},\ \measure[\propfunc]_{0, \zval}\big),
\]
where $d(\mu, \mu')$ is a measure of dissimilarity between probability distributions.
\end{definition}

\paragraph{Subgroup vulnerability from distributional generalization}
In order to extend the worst-case analysis to subgroups, we use the worst-case subgroup vulnerability under adversary's features $\obs$ to the corresponding Bayes adversary: $\vulnz[\obs]$. 
We show that this worst-case subgroup vulnerability is also related to distributional generalization:
\begin{proposition}\label{thm:subgroup-vuln}
    The worst-case vulnerability of a subgroup $\zval$ is bounded:
    \begin{equation}\label{eq:subgroup-vuln}
         \vulnz[\obs] \leq \tvoverfitz{\zval} 
    \end{equation}
    Moreover, for subgroup-aware adversaries the bound becomes an equality:
    \begin{equation}\label{eq:subgroup-vuln-disc}
         \vulnz[\obs,\z] = \tvoverfitz{\zval}
    \end{equation}
\end{proposition}

We defer the proof to \cref{sec:proofs}.

\paragraph{Formalizing disparate vulnerability}
Finally, having discussed subgroup vulnerability, we can analyze disparate vulnerability. Formally, let us define \newterm{disparity in vulnerability}.
\begin{definition}
    Disparity in vulnerability (or disparity for short)
    between two subgroups $\zval$ and $\zval'$ is the difference in vulnerability of
    these subgroups:
    \[\deltavuln[\obs] \define \vulnz[\obs] - \vulnzprim[\obs] \,.\]
\end{definition}

The previous results on the connection between subgroup vulnerability and distributional generalization enable us to relate disparity to degrees of distributional generalization across different population subgroups. From \cref{thm:subgroup-vuln}, we can see that the magnitude of disparity can be trivially bounded using distributional-generalization gaps of the involved subgroups:
\begin{corollary}\label{thm:dv-upper-bound}
    Magnitude of disparity between subgroup $\zval$ and $\zval'$ is upper bounded:
    \begin{equation}\label{eq:dv-upper-bound}
        \big| \deltavuln[\obs] \big| \leq \max\{\tvoverfitz{\zval}, \tvoverfitz{\zval'}\}
    \end{equation}
\end{corollary}

Moreover, disparity has an exact closed form for subgroup-aware adversaries:
\begin{corollary}\label{thm:ns-condition-dv}
    Suppose that a subgroup-aware adversary uses features $(\obs, \z)$. Then, disparity
    between subgroups $\zval$ and $\zval'$ is the difference between distributional generalization
    gaps of these subgroups:
    \begin{equation}\label{eq:ns-condition-dv-disc}
        \begin{aligned}
        \deltavuln[\obs,\z] &= \tvoverfitz{\zval} - \tvoverfitz{\zval'}\,.
        \end{aligned}
    \end{equation}
\end{corollary}

\subsection{Takeaways}
\label{sec:vuln-takeaways}
\paragraph{Necessary and sufficient condition for MIA existence} Without making any parametric assumptions, we have showed that the vulnerability to MIAs can be characterized using an extended notion of generalization, and that disparity is bounded by the difference in levels of distributional generalization across population subgroups.
This interpretation of a standard result in statistical theory generalizes and complements the
theoretical findings of \citet{YeomGFJ18} and \citet{SablayrollesDSO19}. It also confirms that the presence of standard overfitting is not a necessary condition for MIAs to succeed~\cite{YeomGFJ18, LongBWBWTGC18}. 

\paragraph{Hardness of defending against MIAs} The interpretation of worst-case vulnerability through distributional generalization has important consequences for practical defences against MIA that do not rely on differential privacy. 

In order to reduce the vulnerability against adversaries that use features $\obs$, the distribution of $\obs$ for examples that are outside of the training set has to be close to that for the training set examples.
This means that, to avoid vulnerability, a target model has to---either implicitly or explicitly---learn the distribution of $\obs$~\cite{KearnsMRRSS94}  which is a stronger requirement than what is typically necessary for its main task (i.e. generalization in terms of accuracy, or average error).

Moreover, adversaries are not limited to one set of features $\obs$; thus, the distribution has to be learned for a multitude of possible configurations of adversarial features $\obs$.
Additionally, to prevent disparity in vulnerability, the distribution of $\obs$ has to be learned across population subgroups, which is an even more challenging task.
\section{Detecting and Measuring Disparate Vulnerability}
\label{sec:measuring}

We showed in \cref{sec:why-mia} that vulnerability to MIAs appears when a model lacks in distributional generalization. The degree to which records are vulnerable can vary across subgroups in the data, potentially resulting in disparate vulnerability. In this section, we provide mechanisms to reliably estimate subgroup vulnerability and its disparity in practice. 

To empirically estimate MIA vulnerability, we simulate the $\miagame$ game with a real attack. If we could play the game infinite times, then estimating the success probability of the adversary would be trivial. In practice, however, we can only run the game a finite amount of times, which provides us with a finite number of challenge examples $\xval$. We group these examples into two sets of datasets of $n$ elements: a set of $r$ datasets $\{ \inset_i \}_{i = 1 .. r}$ composed of $n$ ``in'' examples (i.e., sampled as in line 4 of the $\miagame$ game, used for training), and $r$ datasets $\{ \outset_i \}_{i = 1 .. r}$ composed of $n$ ``out'' examples (i.e., sampled as in line 6 of the $\miagame$ game, not used for training). Each pair of datasets $\inset_i$ and $\outset_i$ can be seen as the train and test datasets of one model. 

We define the estimate of vulnerability as:
\begin{equation}
   \estvulnadv[(\adv)] \define \frac{1}{r} \sum_{i = 1}^r v_{i}
\end{equation}
where $v_i$ is the \emph{model-specific estimate of vulnerability}: the advantage of the adversary against a single target model. We compute $v_i$ for a pair of datasets $\inset_i$ and $\outset_i$ as:
\begin{equation}\label{eq:est-vuln-model}
    \begin{aligned}
    v_{i} & \define 2 \cdot \frac{1}{2n} \Bigg( \sum_{j = 1}^{n} \id[\adv(\inset_{i}^{(j)}, \clf[\inset_i], n, \datagen) = 1] \\
             & + \sum_{j = 1}^{n} \id[\adv(\outset_{i}^{(j)}, \clf[\inset_i], n, \datagen) = 0]\Bigg) - 1,
    \end{aligned}
\end{equation}
As $r$ increases, $\estvulnadv[(\adv)] $ approximates the value of the true vulnerability $\vulnadv[]$.

We use the same approach to estimate subgroup vulnerability $\vulnadvz[(\adv)]$, but we only use examples that belong to the subgroup of interest $\zval$ when computing the model-specific estimate of subgroup vulnerability $v_{i, \zval}$. We omit $\adv$ when it is clear from context.

\subsection{Statistical Detection of Disparity}
\label{sec:stats}
When evaluating subgroup vulnerability, we have to rely on subsets of $(\inset_i, \outset_i)$ formed by subgroup examples. These subsets are possibly of size much smaller than $n$. Due to the variance of the empirical averages in the \cref{eq:est-vuln-model}, an estimate of subgroup vulnerability is in general less statistically reliable than the estimate of overall vulnerability that uses datasets $(\inset_i, \outset_i)$ in their entirety. As a result, when estimating disparate vulnerability using the estimates of subgroup vulnerability, we need to statistically ensure that, if found, disparity is not due to random chance.

More formally, given estimates $\{v_{i, \zval}\}_{i = 1..r}$ across different subgroups, we want to find statistical evidence that the actual subgroup vulnerabilities differ:
\begin{equation}\label{eq:disparity-test}
    \vulnadv_{\zval_1} \stackrel{?}{\neq} \vulnadv_{\zval_2} \stackrel{?}{\neq} \ldots \stackrel{?}{\neq} \vulnadv_{\zval_t}
\end{equation}

\paragraph{Multiple subgroups}
This problem is an instance of a standard within-subjects experimental design: We have multiple measurements (model-specific vulnerability estimates for different subgroups $v_{i, \zval_1}, v_{i, \zval_2}, \ldots, v_{i, \zval_t}$) for the same subject (model $\clf[\inset_i]$). We want to know whether the means of vulnerability values differ across subgroups. Therefore, we can determine whether the training algorithm exhibits disparate vulnerability using the repeated-measures one-way \textsc{anova} model (see, e.g., \citet[Chapter~14]{Seltman12}).
This approach enables us to use the \textsc{anova} F-test to establish whether there is evidence of disparate vulnerability. Following the standard protocol, if the F-test is positive, we perform \newterm{post-hoc followup tests} to determine which particular pairs of subgroups exhibit disparity. For the post-hoc tests, we use pairwise dependent t-tests with correction for multiple comparisons. As the correction method, we use the standard Benjamini-Hochberg procedure for controlling the false detection rate.

\paragraph{Two subgroups} When comparing only two subgroups, $\zval$ and $\zval'$, the procedure naturally simplifies to running one dependent t-test that checks if the difference between means of two groups is significant. 

\subsection{The Bias Problem}
\label{sec:bias-eval}
Some attacks in the literature assume that the adversary has \newterm{additional knowledge} beyond the tuple $(\xval, \clf[\inset], n, \datagen)$. This knowledge can result in the vulnerability estimation being positively biased: indicating higher vulnerability than the actual worst case within the knowledge model of $(\xval, \clf[\inset], n, \datagen)$. Overestimating vulnerability is not necessarily an issue, as pessimistic estimates incentivize caution in deployment. However, if the positive bias is correlated with the parameters of a subgroup (e.g., higher bias for smaller subgroups), it leads to incorrect conclusions about \emph{disparate} vulnerability.

In this section, we check whether estimates of vulnerability using attacks proposed in the literature are biased.
We evaluate three attacks:
\begin{itemize}
    \item \textbf{Shadow-model attack~\cite{ShokriSSS17}}. An adversary trains a number of shadow models using the target training algorithm $\train(\cdot)$ on datasets sampled from $\datagen^n$. The adversary uses these shadow models to train a machine-learning classifier to estimate the probability $\Pr[\ismember \mid \obs]$. In our evaluation, we use 30 shadows and Gradient Boosting Trees as the attack classifier.
    \item \textbf{Average-threshold attack~\cite{YeomGFJ18}}. An adversary has additional knowledge: the average loss on the training dataset $\tau$ and the loss function $\loss$ used to compute this average, $\big(\tau,\ \loss(\cdot, \cdot) \big)$, where $\tau \define \sum_{\xval \in \inset} \loss(\clf, \xval)$. When attacking, the adversary uses $\tau$ as threshold to decide whether the challenge example was ``in'' (the example's loss less than threshold) or ``out'' (greater than threshold).
    \item \textbf{Optimal-threshold attack~\cite{ShokriSZ19, ChangS21}}. An adversary has additional knowledge: the loss function $\loss$ and the optimal loss threshold $\tau^*$ that separates the losses in the best way, $\big(\tau^*,\ \loss(\cdot, \cdot))$, where
    \[
        \begin{aligned}
            \tau^* \define \arg \max_{\tau} & \frac{1}{n} \sum_{\xval \in \inset} \id[ \loss(\clf, \xval) \leq \tau] \\
                   & + \E_{\xval \sim \datagen}\big[\id[\loss(\clf, \xval) > \tau]\big]
        \end{aligned}
    \]
    The attack proceeds as the average-threshold attack.
\end{itemize}

We deviate slightly from the attacks' original formulations. The threshold attacks use $\obs = \loss(\clf, \x)$ as features, where the loss function is cross-entropy, whereas the original shadow-model attack used $\obs = (\yhat, \y)$. For fairness, we make all adversaries use the threshold attacks' features.

As we want to evaluate subgroup-aware adversaries, we use features $\obs = \big(\loss(\clf, \x), \z\big)$ for all attacks, with cross-entropy as loss function. We make the attacks subgroup-aware as follows. For the shadow-model attack, the adversary trains separate attack classifiers for each subgroup, and then applies the appropriate classifier to each challenge example. For the threshold attacks, we assume the adversary has different thresholds for each subgroup \cite{ChangS21, SongMittal21}, i.e., average loss, respectively optimal threshold, per subgroup.

\paragraph{Method} It is hard to tell exactly if an estimate is higher than the worst-case vulnerability, as in practice the worst case is unknowable. We propose a simple test for bias within our adversarial model: run the estimation method against \newterm{data-independent models}. A target model can be independent of its training data, e.g., if it is completely random, constant, or trained with differential privacy parameter $\varepsilon \approx 0$ (see \cref{sec:dp}). If the model is independent of the data, we expect the estimates of overall and subgroup vulnerabilities, as well as disparity, to all be zero in expectation. We refer to any violation of this property as \newterm{null-model bias}. We are not only interested in whether a method exhibits such bias, but in whether this bias is correlated with subgroups.

\begin{figure}
    \centering
    \includegraphics[width=\linewidth]{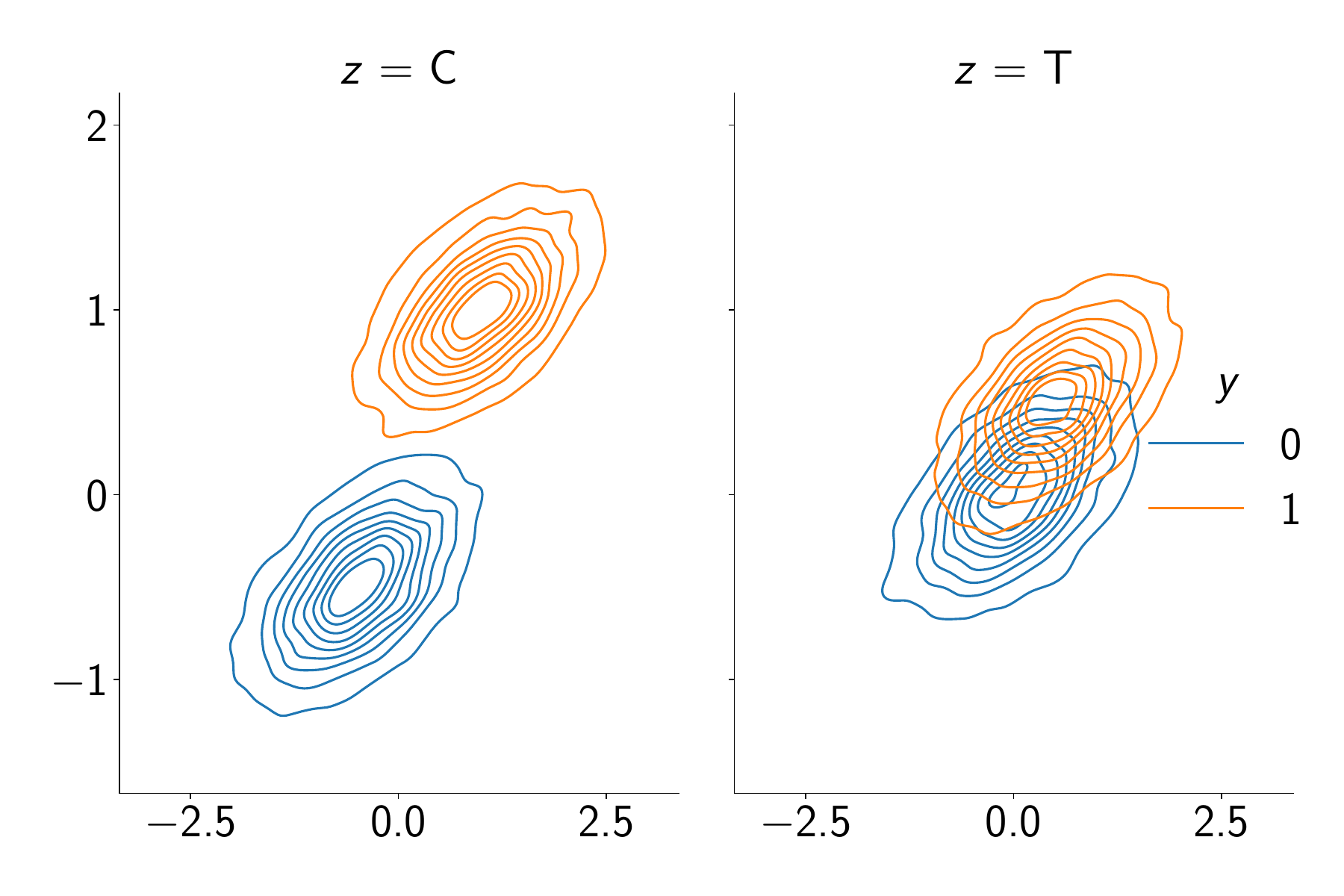}
    \caption{Distribution of values in our synthetic data. \emph{x-axis:} value of the 1-st dimension of the synthetic data, \emph{y-axis:} value of the 2-nd dimension. We use 100-dimensional data for our experiments.}
    \label{fig:synthetic-data-demo}
\end{figure}

\paragraph{Dataset} To have control over the distributions of subgroups and their representation, we create a synthetic dataset. We assume that the examples have binary class labels $\yval \in \{0, 1\}$, and belong to one of two subgroups $\zval \in \{C, T\}$. We generate the examples from the multivariate normal distributions:

\[
    \begin{aligned}
        \Pr(\xval \mid \yval=0, \zval=C) & \sim \mathcal{N}(-\nicefrac{1}{2} \cdot \mathbf{1}^d, \Sigma) \\
        \Pr(\xval \mid \yval=1, \zval=C) & \sim \mathcal{N}(1 \cdot \mathbf{1}^d, \Sigma) \\
        \Pr(\xval \mid \yval=0, \zval=T) & \sim \mathcal{N}(0 \cdot \mathbf{1}^d, \Sigma) \\
        \Pr(\xval \mid \yval=1, \zval=T) & \sim \mathcal{N}(\nicefrac{1}{2} \cdot \mathbf{1}^d, \Sigma),
    \end{aligned}
\]
where $\mathbf{1}^d$ is a $d$-vector of all ones, and the covariance matrix $\Sigma$ is generated such that $||\Sigma||_{\max} \leq 1$. We use $d = 100$ dimensions, and set $\Pr[\yval = 1] = \nicefrac{1}{2}$.  See \cref{fig:synthetic-data-demo} for an illustration.

To reflect that some subgroups can be harder to learn than others, the distributions are designed in such a way that the subgroup $\zval = C$ is more separable and hence more easily learnable than the subgroup $\zval = T$. In our experiments we use the subgroup $\zval = C$ as the control (or \emph{majority}) subgroup with fixed number of representatives in the data, and $\zval = T$ as the treatment (or \emph{minority}) subgroup whose size we vary.

\paragraph{Setup} 
To see if the potential null-model bias depends on the sizes of subgroups, we generate multiple synthetic datasets such that each contains data belonging to two subgroups: control and treatment. The control subgroup has 1000 representatives in each dataset; the size of the treatment subgroup varies between 25 and 1000, with 8 distinct values. We run 8 experiments with different subgroup proportions. Within each experiment, we train 200 target models on freshly generated datasets. We set the target training algorithm to output the same classifier for any input training dataset. Recall that because the models are independent of the input, we expect all vulnerability estimates to be zero on average. We estimate disparity using three attacks described above, and run t-tests to see if the estimates are statistically significant as explained in \cref{sec:stats}.

\begin{figure}[t]
    \centering
    \includegraphics[width=\linewidth]{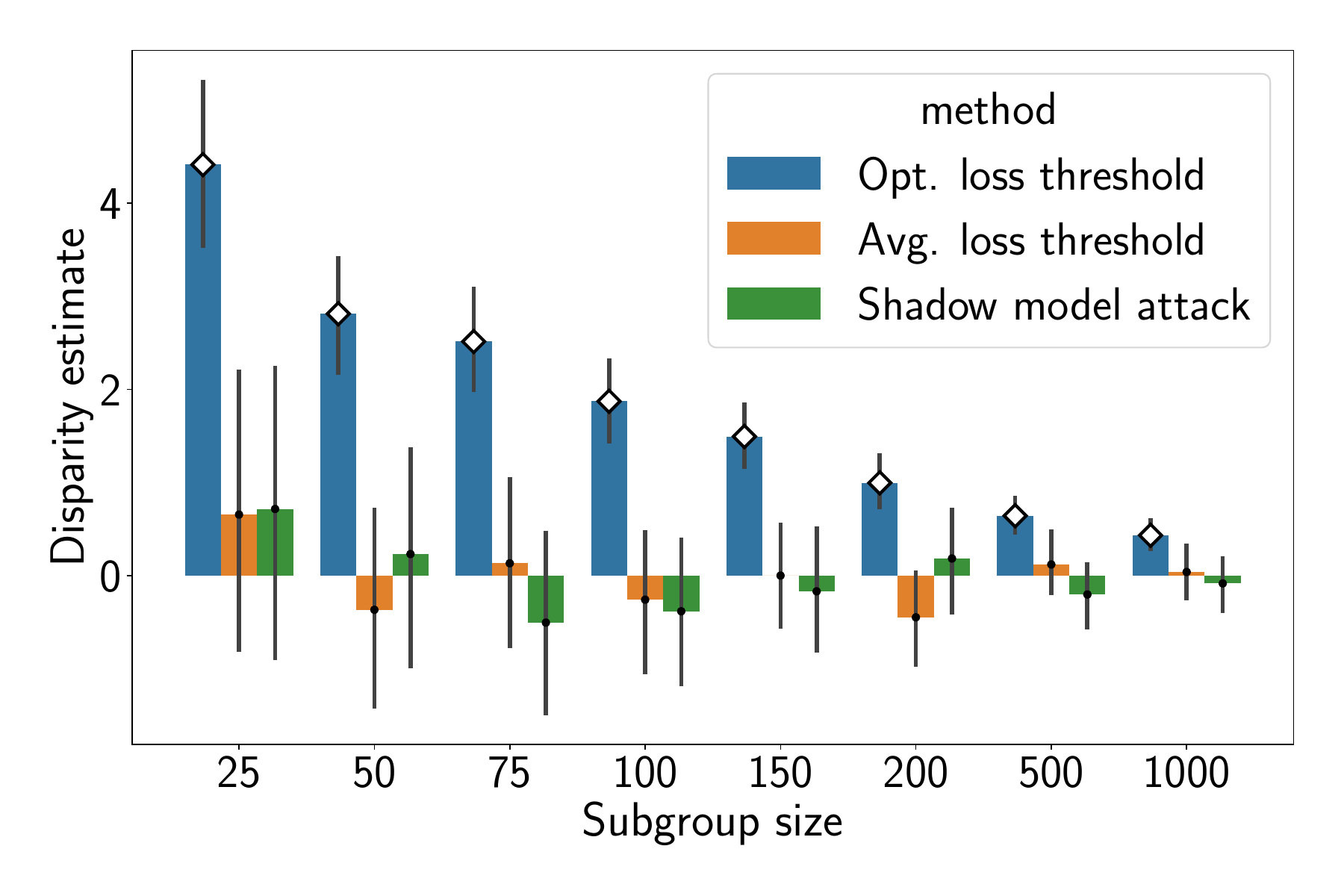}
    \caption{Null-model bias of methods to estimate disparate vulnerability. Disparity in percentage points (\emph{y-axis}) vs. size of the treatment subgroup in the training data (\emph{x-axis}).
    Computed on synthetic datasets with fixed control subgroup (1000 examples) .
    The target training algorithm is data-independent: actual MIA vulnerability, subgroup vulnerabilities, and disparity in vulnerability are all zero. The error bars represent the variation across 200 model-specific estimates. The diamond marker ($\lozenge$) means that an estimate significantly differs from zero with $p < 0.001$.}
    \label{fig:bias-experiments}
\end{figure}

\paragraph{Results on our synthetic dataset} In \cref{fig:bias-experiments}, we can see that the estimates of disparity produced with the shadow-model attack and the average-threshold attack are centered around zero, with the statistical tests confirming no significant difference from zero. The estimates coming from the optimal-threshold attack, however, are highly biased compared to the other attacks, as the estimates are consistently and significantly ($p < 0.001$) different from zero. The bias is always positive --- overestimates disparity --- and gets higher as the size of the treatment subgroup decreases. As the target models are independent of their training data and thus cannot have disparate vulnerability, we conclude that the use of the optimal-threshold attack results in significant null-model bias that grows as the subgroup size gets smaller.

\paragraph{Results on the dataset by \citet{ChangS21}}
To verify that our results are not artifacts of our specific synthetic data setup, we also reproduce the data setup used by \citeauthor{ChangS21} to evaluate their subgroup-aware optimal-threshold attack. In their setup, they have one fixed dataset containing four subgroups that we denote as ``0-0'', ``0-1'', ``1-0'', ``1-1'', where the first number indicates simulated demographic group and the second number the class $\yval$ (we refer to the original work~\cite{ChangS21} for details). The subgroups have 50, 450, 1000, and 1000 examples, respectively, with the total dataset size of 2500 examples. Following \citeauthor{ChangS21}, we randomly subsample training datasets of size 1250 from the full dataset, and train one model on each. As before, we ``train'' a data-independent model. In this experiment, we only use threshold attacks due to the small size of the dataset (see \cref{sec:realworld} for more details). We use the \textsc{anova} F-test as described in \cref{sec:stats} to determine whether any of the subgroups have differing subgroup vulnerabilities.

\begin{figure}[t]
    \centering
    \includegraphics[width=\linewidth]{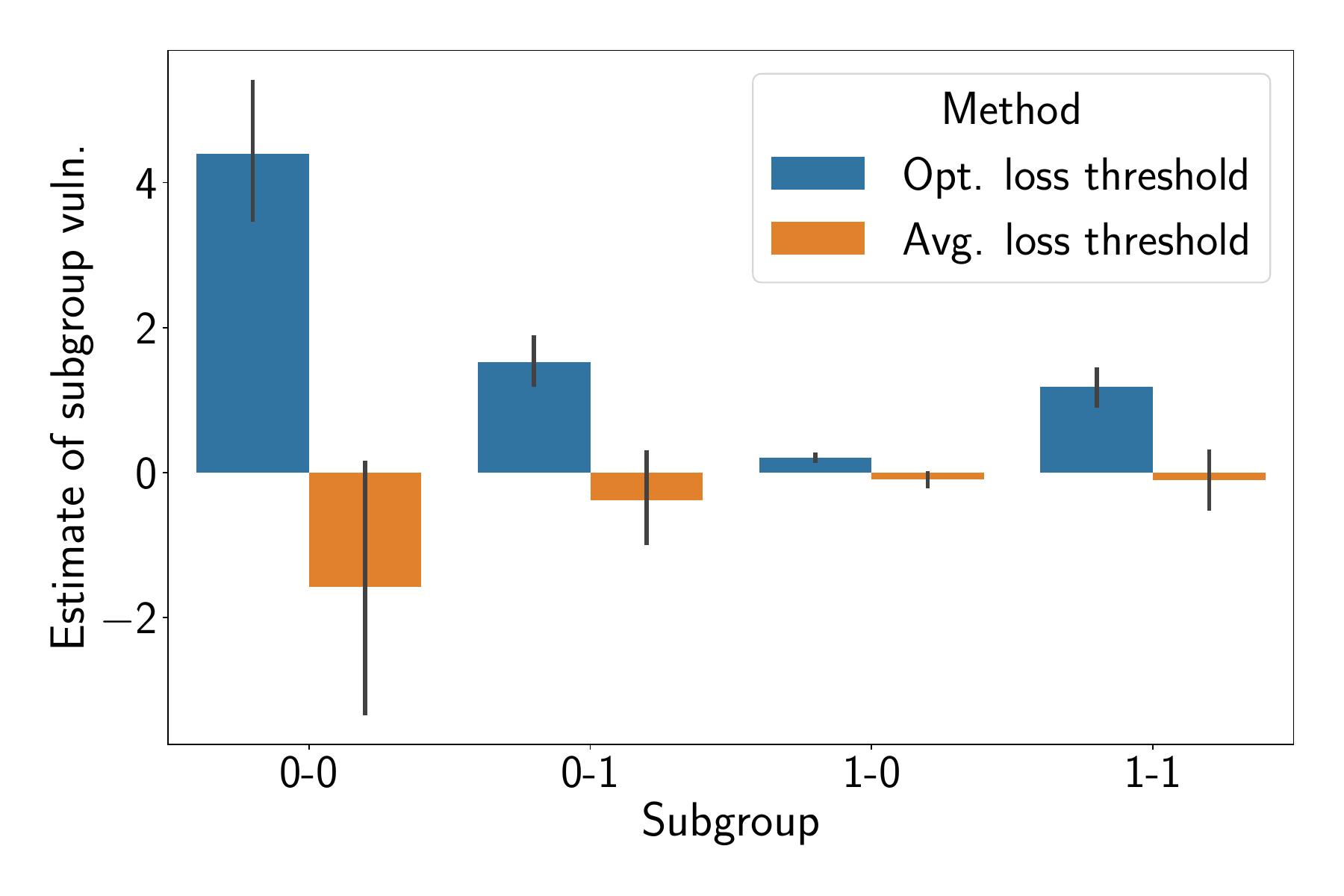}
    \caption{Null-model bias on the synthetic data setup from \citet{ChangS21}. Estimate of disparity in percentage points (\emph{y-axis}) vs. subgroup (\emph{x-axis}). The target training algorithm is data-independent, thus actual MIA vulnerability, subgroup vulnerabilities, and disparity in vulnerability are all zero.}
    \label{fig:bias-experiments-chang}
\end{figure}

\cref{fig:bias-experiments-chang} shows that significant null-model bias of the optimal-threshold attack also appears on this dataset (F-test $p < 0.001$). In particular, the subgroup vulnerability for the smallest subgroup ``0-0'' with 50 examples appears as 4\%.  At the same time, the estimates from the average-threshold attack are centered around 0 and do not significantly differ (F-test $p \approx 0.1$), suggesting no null-model bias.

This bias, however, should not affect the conclusions by \citet{ChangS21}. Rather than directly using the estimates of subgroup vulnerability, their analysis used \emph{differences} in estimates of subgroup vulnerability between two models (a ``fair'' and a ``regular'' model). 
In their particular scenario, the bias introduced by the estimation should be cancelled out in the final difference. Although the conclusions of \citeauthor{ChangS21} should not be affected by the bias, estimation methods such as the optimal-threshold attack should be avoided when evaluating disparate vulnerability in general.

\paragraph{Biased estimator in a prior version} A pre-print version of our work\footnote{\href{arxiv.org/abs/1906.00389v2}{https://arxiv.org/abs/1906.00389v2}} used a vulnerability estimation method that, like the optimal-threshold attack, leveraged information about the training dataset of the target model. This estimator was therefore biased, and so were the numerical results of that version.

\paragraph{Takeaways} Biased estimators of vulnerability can result in consistent overestimation of disparity if the bias correlates with subgroup parameters. The shadow-model attack does not have such bias as it does not have access to any information about a specific target. Interestingly, the average-threshold attack, despite using an additional piece of knowledge that goes beyond our adversarial model, also does not exhibit such bias. On the contrary, the optimal-threshold attack produces significantly biased estimates for small groups. 

Our results show the need to evaluate bias of the estimation method when measuring disparate vulnerability. To this end, we proposed to measure null-model bias, which detects bias when the worst-case vulnerability is zero. This test does not preclude a method from having bias if the worst-case vulnerability is larger. However, in practice MIA vulnerability has been shown to be relatively low.

\subsection{Does Disparate Vulnerability Exist in ML Models?}
\label{sec:experiments-main}
Having established suitable methods for measuring disparate vulnerability, we apply them in a synthetic setup, and show that disparate vulnerability does arise in practice. 

\begin{figure}
    \includegraphics[width=\linewidth]{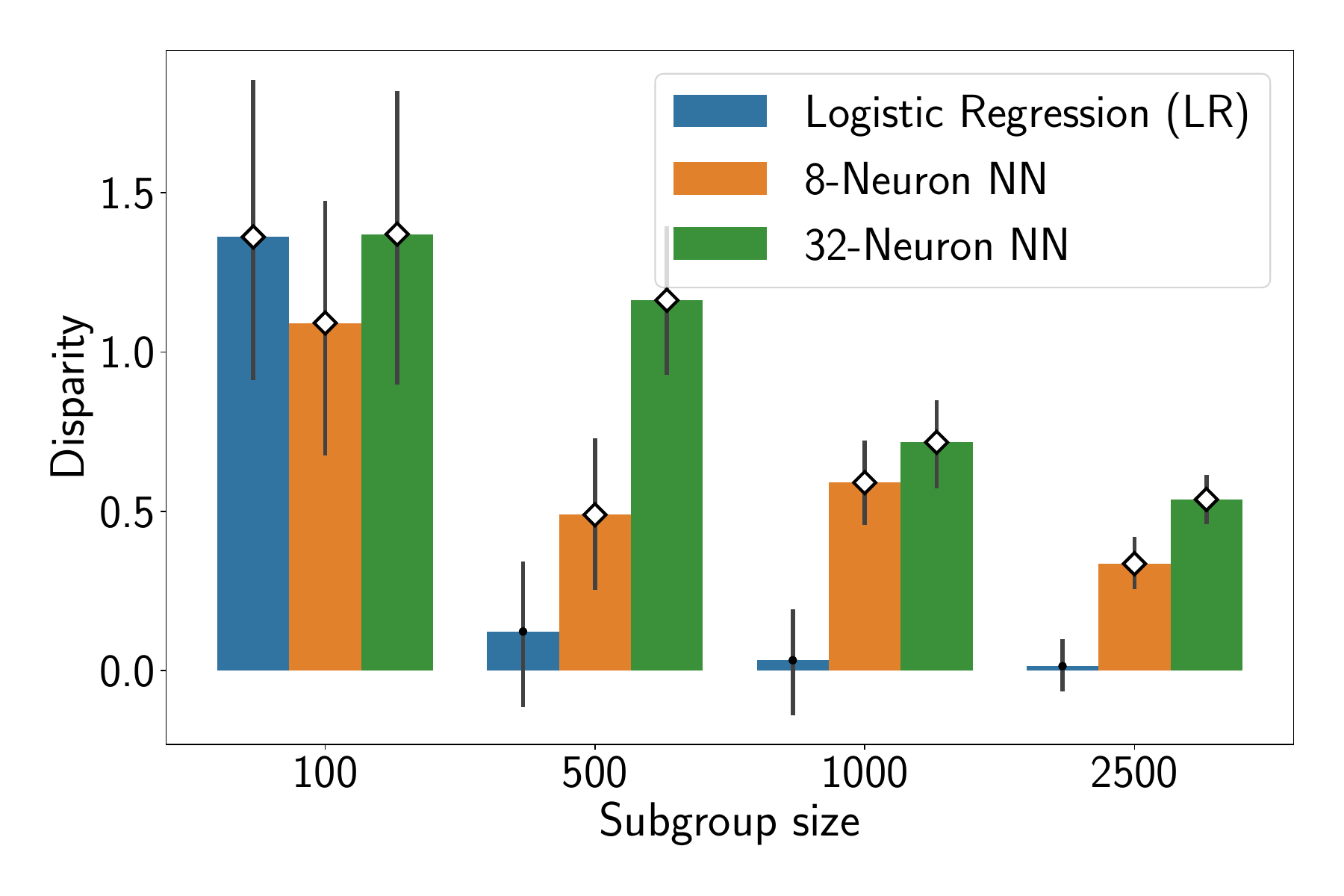}
    \caption{Disparate vulnerability vs. subgroup representation in a training dataset. The \textit{y-axis} represents disparity in vulnerability between the treatment group $\zval$ and control group $\zval'$ whose size is fixed to 2500, in percentage points. The error bars represent the variation across 200 model-specific estimates. Statistical significance markers: $p < 0.001$ ($\lozenge$), $p < 0.01$ ($\circ$), $p \geq 0.01$ ($\cdot$).}
    \label{fig:sample-size-normal}
\end{figure}

\paragraph{Setup} 
To capture the effect of subgroup size in the training data, we create several experiments with different subgroup proportions. Within each experiment, we sample 200 dataset pairs $\inset_i$ and $\outset_i$ from our data distribution. In each dataset, the size of the control subgroup is fixed at 2500, and we vary the size of the treatment subgroup between experiments: 100, 500, 1000, and 2500. We estimate subgroup vulnerabilities using the subgroup-aware shadow-model attack (see \cref{sec:bias-eval}), because this attack is guaranteed to not have null-model bias. As before, we use $\obs = (\ell(\clf, \x), \z)$ as adversary's features. To train shadow models, we independently sample 30 fresh datasets from our data distribution. We use t-tests to determine whether measured disparity is statistically significant as described in \cref{sec:stats}.

\paragraph{Targets} We evaluate the following model families: logistic regression, and two ReLU neural networks with one hidden layer containing 8 and 32 neurons, respectively. We use the \emph{scikit-learn} library~\cite{scikit-learn} to train these models. All our models attain close to 100\% test accuracy in our synthetic data setup.

\paragraph{Results} The results in \cref{fig:sample-size-normal} show  that ML models can exhibit disparate vulnerability, even on a simple dataset. For all treatment sizes and targets, our estimates of disparity are significant ($p < 0.001$), with the exception of the logistic regression when the treatment subgroup is relatively well-represented (500 -- 2500 examples). 
We also see that the sample size of the subgroup plays an important role in disparate vulnerability: \emph{the less represented is a group in the training data, the higher the disparate vulnerability as compared to a better represented group.} Even though the sample size seems to be the dominant effect, we observe small but significant disparate vulnerability even when the subgroups are equally represented in training.

\section{Mitigating Disparate Vulnerability}
We now study whether some existing methods for addressing privacy and fairness in ML prevent disparate vulnerability.

\subsection{Fairness Constraints}
\label{sec:fairness}
Due to the dependency of disparate vulnerability on the disparate \emph{behavior} of the
model across subgroups, minimizing the between-subgroup discrepancy in any given property, such as model's outputs or loss~\cite{ChouldechovaR18}, intuitively 
could decrease disparate vulnerability.

\begin{figure*}[t]
    \centering
    \includegraphics[width=0.45\linewidth]{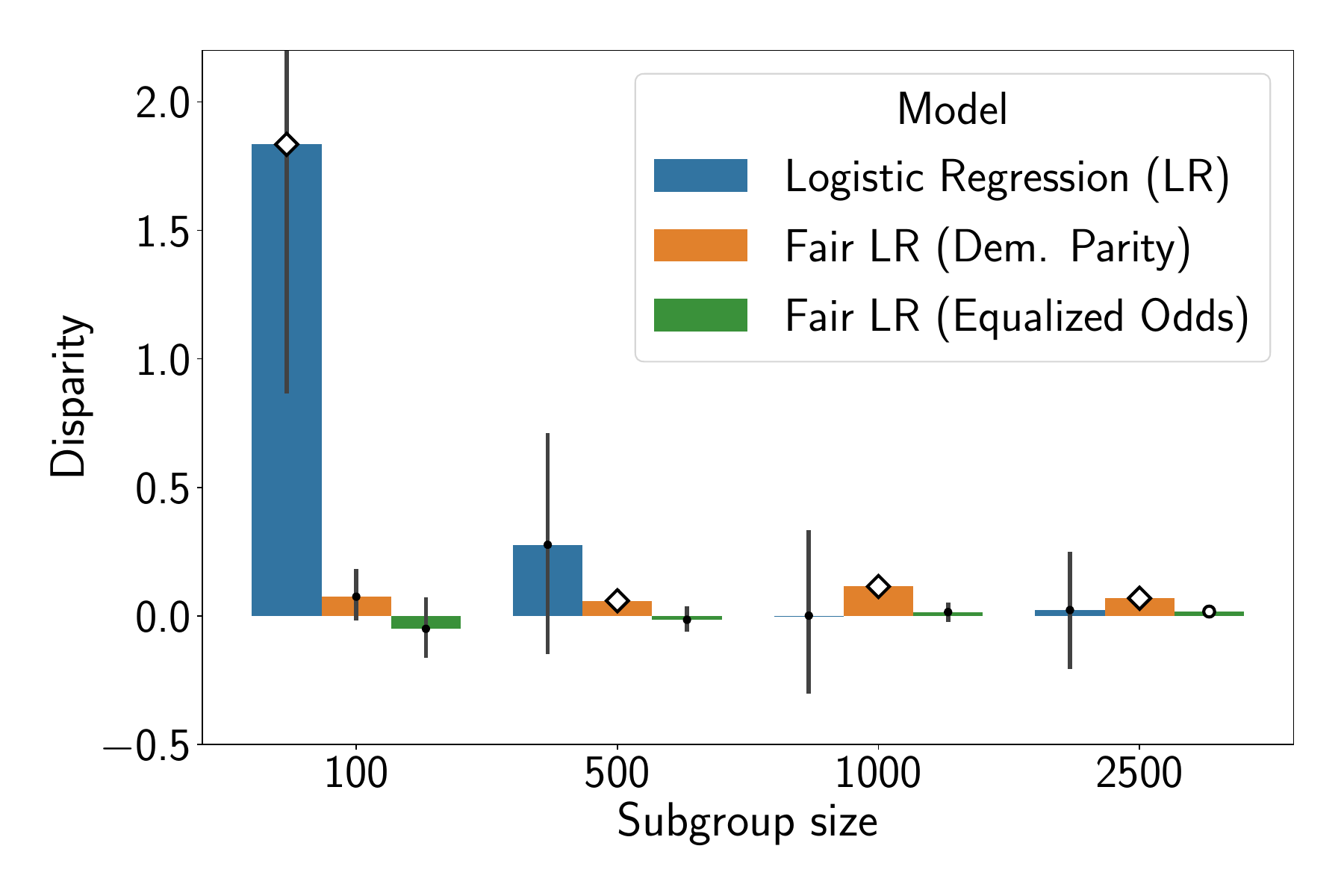}
    \includegraphics[width=0.45\linewidth]{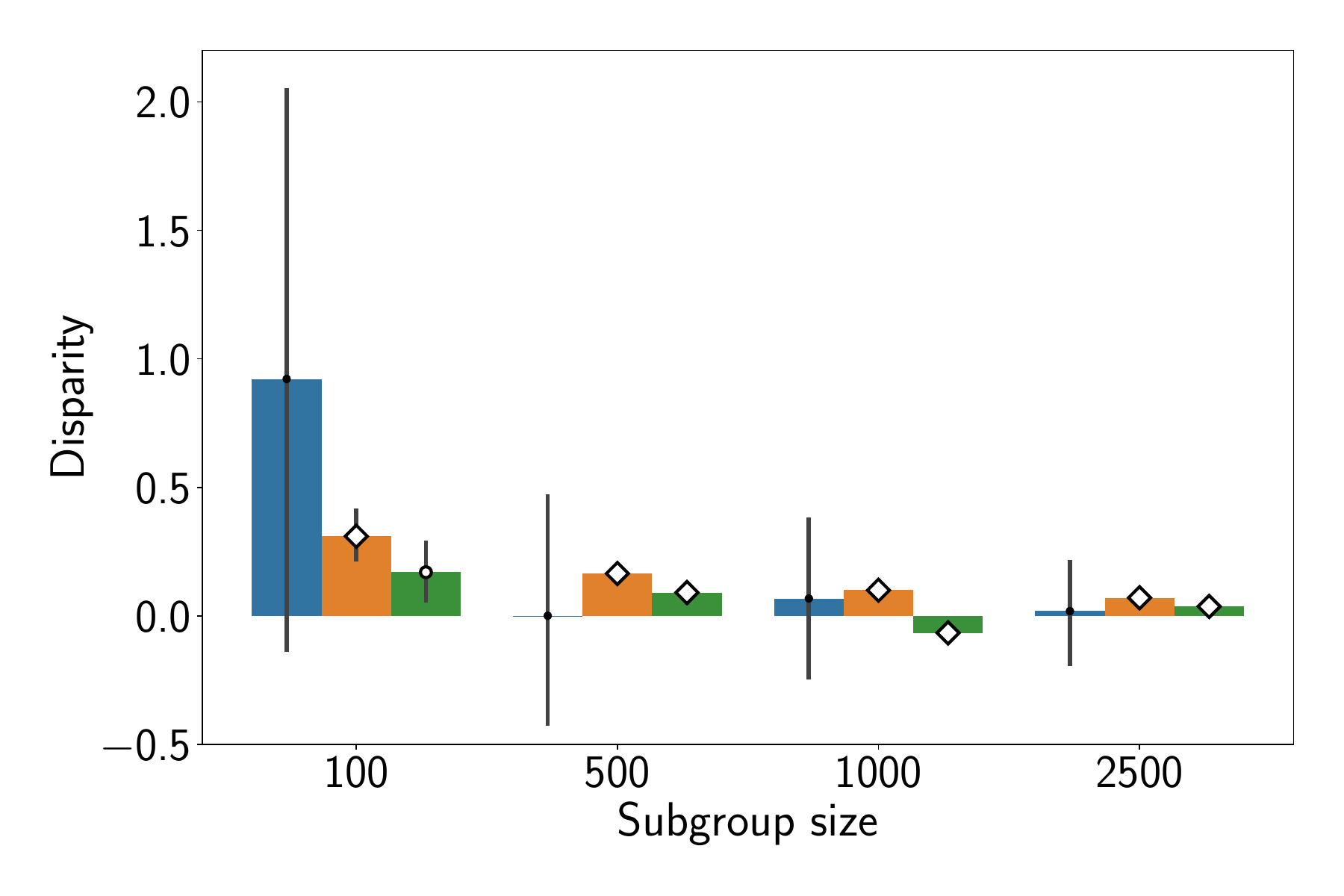}
    \caption{Effect of algorithmic-fairness constraints on disparate vulnerability. The vulnerability is estimated with subgroup-aware attacks that use models' outputs as the feature \emph{(left)}, and the models' loss \emph{(right)}. The results for logistic regression are provided for reference (its values here are not comparable with the results of other experiments as the data dimensionality is different). See \cref{fig:sample-size-normal} caption for details.}
    \label{fig:sample-size-fair}
\end{figure*}

\begin{figure*}
    \centering
    \includegraphics[width=0.45\linewidth]{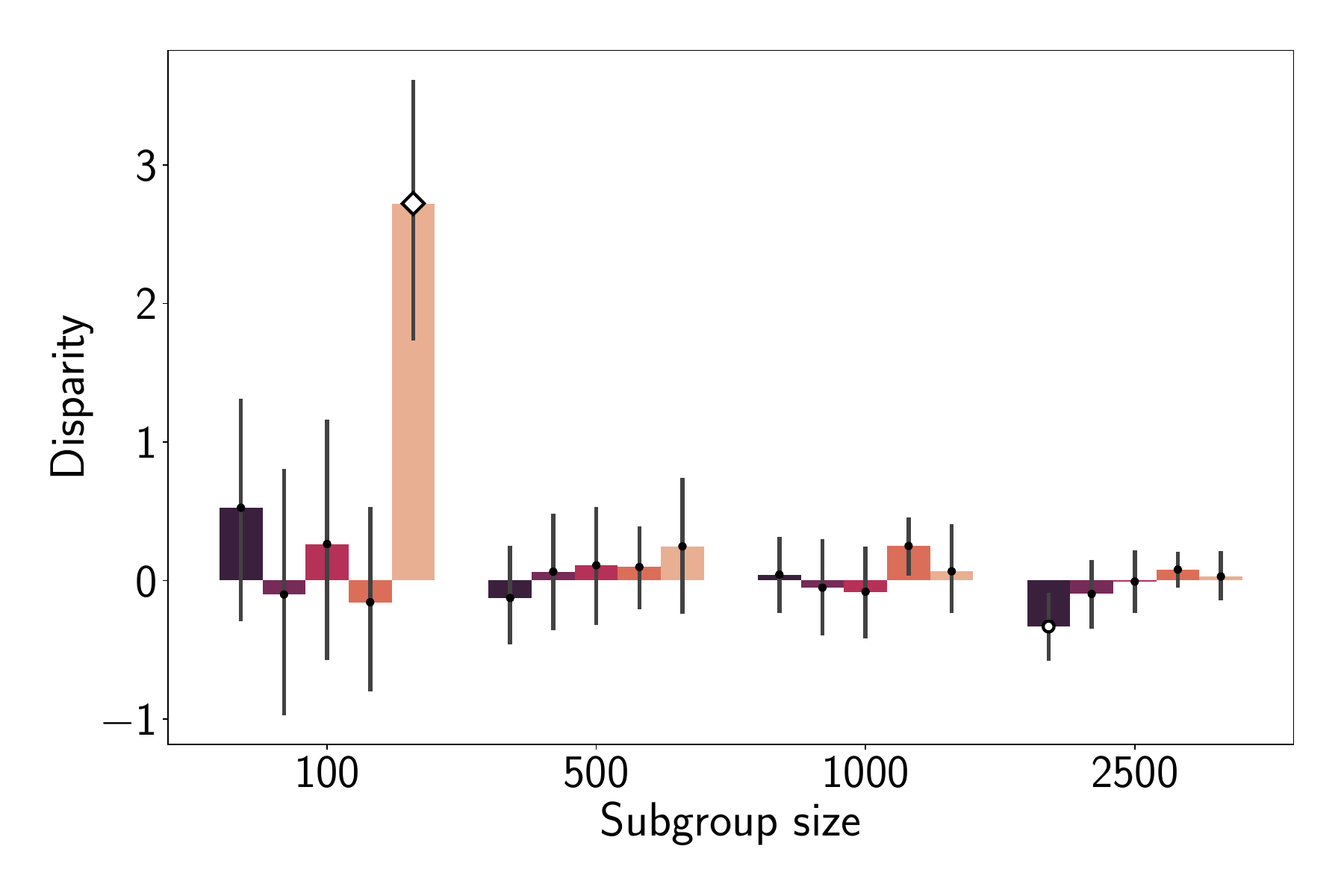}~
    \includegraphics[width=0.45\linewidth]{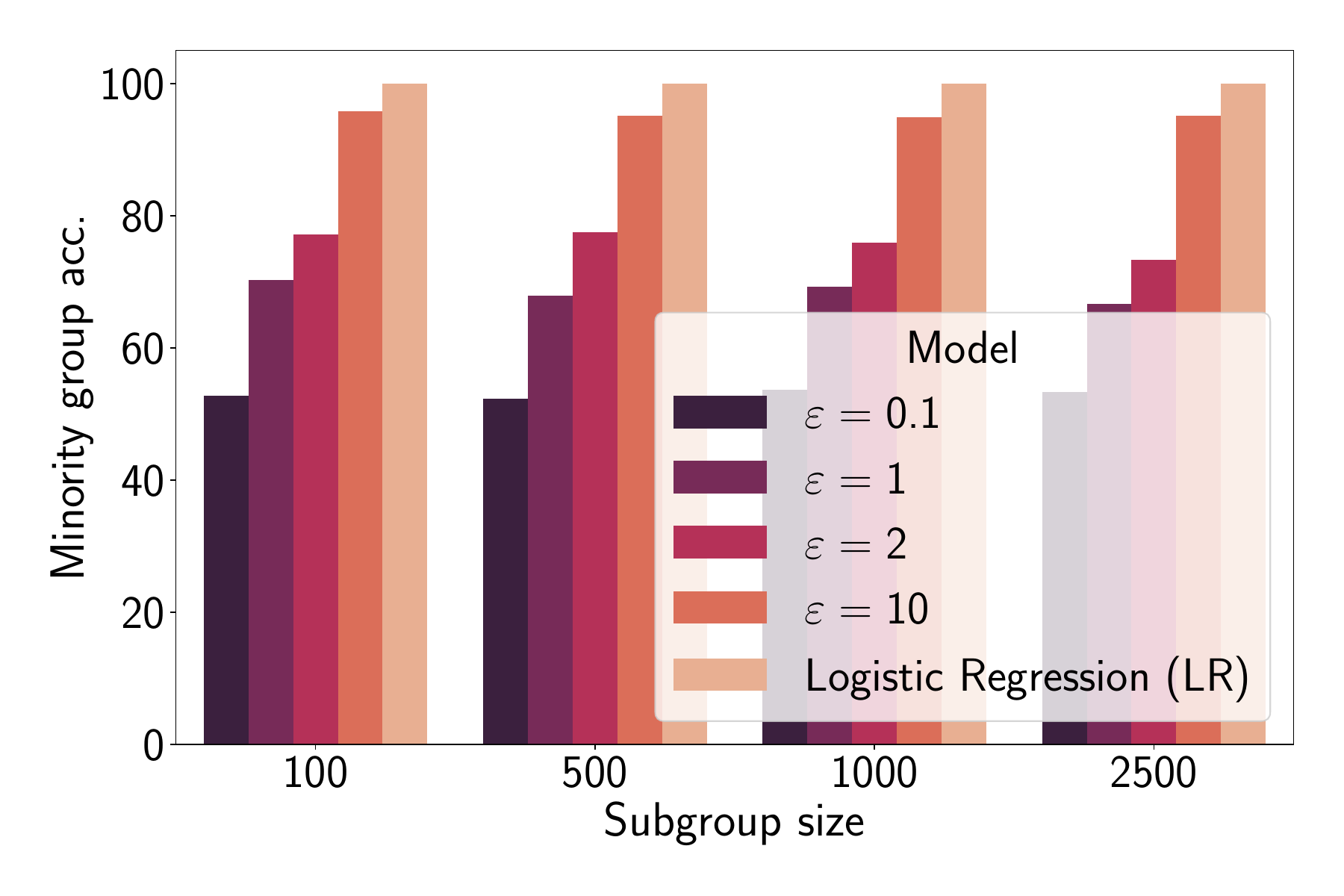}~
    \caption{Effect of differentially private training on disparate vulnerability \emph{(left)}, and test accuracy \emph{(right)}. The results for logistic regression are provided for reference. See \cref{fig:sample-size-normal} caption for details.
    }
    \label{fig:sample-size-dp}
\end{figure*}

Formally, let us denote by $\gap^\propfunc$ the total-variation distance between distributions of some property of a model $\propfunc(\clf, \xval)$ on examples coming from two subgroups $\zval$ and $\zval'$:
\[
    \gap^{\propfunc} \define \TV\left(\Pr_\dtez[\propfunc(\clf, \xval)], \Pr_\dtezprim[\propfunc(\clf, \xval)]\right)
\]
With an appropriate choice of the property function, certain notions of algorithmic fairness can be seen as equivalent, or upper bounding, the above gap. For example, if we choose the model property to be its outputs, then for $\propfunc(\clf, \xval) = \clf(\xval)$, we obtain \newterm{demographic parity}~\cite{DworkHPRZ12}. Similarly, for the 0-1 loss property of the model, choosing $\propfunc(\clf, \xval) = \id[\clf(\xval) = \yval(\xval)]$ gives us \newterm{accuracy equality}~\cite{BerkHHJSKR18}. 

In practice, a notion of fairness is satisfied on the training dataset rather than the whole data distribution. To capture this, we define an in-training gap as follows:
\[
    \gap^{\propfunc}_\inset \define \TV\left( \Pr_\dtrz[\propfunc(\clf, \xval)], \Pr_\dtrzprim[\propfunc(\clf, \xval)] \right)
\]

The following proposition establishes that, if the in-training gap is bounded and the model generalizes its fairness condition well, then vulnerability disparity is bounded to adversaries that use the property addressed by the fairness notion:

\begin{proposition}
\label{thm:fairness-bound}
    Suppose a subgroup-aware adversary uses features $(\obs, \z)$, and the following two conditions are satisfied:
    \begin{enumerate}
        \item Fairness on the training data: $\gap^{\knowledge}_\inset \leq \gamma$
        \item Fairness generalization: $|\gap^{\knowledge} - \gap^{\knowledge}_\inset| \leq \delta$
    \end{enumerate}
    Then, the magnitude of disparity in worst-case vulnerability is bounded as follows:
    \[
        |\deltavuln[\obs,\z]| \leq 2\gamma + \delta
    \]
\end{proposition}

\begin{proof}[Proof of \cref{thm:fairness-bound}]
    First, observe that a combination of the two conditions implies:
    \[
        \gap^{\knowledge} = \TV(\measure[]_{0, \zval}, \measure[]_{0, \zval'}) \leq \gamma + \delta
    \]
    By this implication and the triangle property of total variation we have that:
    \[
        \begin{aligned}
            \TV(\measure[]_{0,\zval'}, \measure[]_{1,\zval'}) \leq &\ \underline{\TV(\measure[]_{1,\zval'}, \measure[]_{0,\zval})}
            + \TV(\measure[]_{0,\zval}, \measure[]_{0,\zval'}) \\
            \leq &\ \underline{\TV(\measure[]_{1,\zval'}, \measure[]_{0,\zval})} + \gamma + \delta \\
        \end{aligned}
    \]
    Applying the triangle inequality to the underlined term:
    \[
        \begin{aligned}
            \underline{\TV(\measure[]_{1,\zval'}, \measure[]_{0,\zval})} & \leq \TV(\measure[]_{0,\zval}, \measure[]_{1,\zval})
            + \TV(\measure[]_{1,\zval}, \measure[]_{1,\zval'}) \\
            & \leq \TV(\measure[]_{0,\zval}, \measure[]_{1,\zval}) + \gamma
        \end{aligned}
    \]
    Combining the two,
    \[
        \begin{aligned}
            \TV(\measure[]_{0,\zval'}, \measure[]_{1,\zval'}) - \gamma - \delta & \leq \underline{\TV(\measure[]_{1,\zval'}, \measure[]_{0,\zval})} \\
            & \leq \TV(\measure[]_{0,\zval}, \measure[]_{1,\zval}) + \gamma
        \end{aligned}
    \]
    Implying:
    \[
        \tvoverfitz{\zval'} - \tvoverfitz{\zval} \leq 2 \gamma + \delta
    \]
    If we apply the previous steps analogously we can also obtain:
    \[
        \begin{aligned}
            \TV(\measure[]_{0,\zval}, \measure[]_{1,\zval}) - \gamma - \delta & \leq \TV(\measure[]_{1,\zval}, \measure[]_{0,\zval'}) \\
            & \leq \TV(\measure[]_{0,\zval'}, \measure[]_{1,\zval'}) + \gamma \\
        \end{aligned}
    \]
    Thus,
    \[
        \tvoverfitz{\zval} - \tvoverfitz{\zval'} \leq 2 \gamma + \delta
    \]
    Combining the inequalities, we get:
    \[
        |\tvoverfitz{\zval} - \tvoverfitz{\zval'}| \leq 2\gamma + \delta
    \]
    By \cref{thm:ns-condition-dv}, we obtain the sought bound.
\end{proof}

We note that these guarantees only apply to adversaries targeting the features addressed by implemented the fairness notion. In other words, just as in algorithmic-fairness literature where no single fairness measure is appropriate in a general context~\cite{FriedlerSV16}, no one fairness measure can provide guarantees for bounding disparate vulnerability for any adversary.
\subsubsection{Empirical Evaluation}

\paragraph{Fairness notions} To validate the theoretical results, we estimate vulnerability of models that satisfy two algorithmic-fairness notions: First, \emph{demographic parity}~\cite{DworkHPRZ12} which ensures that distributions of model outputs between demographic subgroups are close: $\gap^{\knowledge[\yhat]} \approx 0$. Second, \newterm{equalized odds}, which ensures that true-positive rates and false-positive rates between the subgroups are close~\cite{HardtPNS16}. We choose these notions as they are common in the literature, and there exist efficient algorithms and tooling for producing classifiers that satisfy them.
To train the classifiers, we use the threshold post-processing approach~\cite{HardtPNS16} from the \textit{fairlearn} library~\cite{fairlearn}, applied to a logistic regression classifier.

\paragraph{Setup} Within the setup of \cref{sec:experiments-main}, we run the following two experiments:

\begin{itemize}
    \item[E1] We fulfill the requirements of \cref{thm:fairness-bound}. For this, we estimate vulnerability using features equalized by demographic parity: $\obs = \left(\yhat, \z\right)$. By \cref{thm:fairness-bound}, we expect low disparity in vulnerability \emph{for both classifiers} as long as they generalize their fairness property well. In \cref{sec:proofs}, we show that in our data setup equalized odds implies demographic parity, thus the theoretical guarantee also applies for equality of odds.
    \item[E2] We estimate vulnerability using adversary's features $\obs = \left( \loss(\clf, \x), \z \right)$ which do \emph{not} match what the fairness property does, so the requirements of \cref{thm:fairness-bound} are not fulfilled. 
\end{itemize}

We find that with 100 dimensions in our data setup, the threshold-optimization algorithm produces models that classify the data with 100\% accuracy and no vulnerability. Thus, to demonstrate a setting where disparate vulnerability arises, we deviate from the parameters of \cref{sec:experiments-main} and we use the synthetic dataset with 10 dimensions.

\paragraph{Results} We present the results in \cref{fig:sample-size-fair}. For E1, we see that demographic parity decreases disparate vulnerability compared to standard logistic regression. This empirically confirms \cref{thm:fairness-bound}.
For E2, as expected, both equalized odds and demographic parity do not completely prevent disparate vulnerability. Yet, they do decrease its magnitude by 3$\times$ compared to the standard logistic regression. 

In our particular setup, the constrained models do not perform worse than the unconstrained models. In general, however, fairness notions can be inherently at odds with accuracy~\cite{ZhaoG19}.

\subsection{Differentially Private Training}
\label{sec:dp}
In this section, we look at how learning with differential privacy \citep{Dwork11} relates to disparity in vulnerability. We use the basic notion of differential privacy:
\begin{definition}
	Training algorithm $\train$ 
    satisfies $\varepsilon$-differential privacy (DP)
	if for any two datasets $\inset, \inset'$ differing by the records of one
    individual, for any set of models $T$:
	\[
        \Pr[\clf[\inset] \in T] \leq \exp(\varepsilon) \Pr[\clf[\inset'] \in T]
    \]
\end{definition}

DP training limits the contribution of any individual in the dataset to the model training. Thus, DP should decrease vulnerability to MIAs.
In particular, \citet{YeomGFJ18}, \citet{ChatzikokolakisCPT20} and \citet{HumphriesRTOGK20}, 
showed the advantage of a MIA adversary is bounded by DP in the setting of the \miagame game. For example:
\begin{proposition}[Adapted from \citet{YeomGFJ18}]\label{thm:dp-bound}
If the training algorithm satisfies $\varepsilon$-DP, the worst-case vulnerability with any adversary's features $\obs$ is bounded:
\begin{equation}\label{eq:dp-bound-overall}
    \vuln[\obs] \leq \exp(\varepsilon) - 1
\end{equation}
\end{proposition}

These guarantees extend to disparate vulnerability:
\begin{proposition}\label{thm:dp-bound-disparity}
If the training algorithm satisfies $\varepsilon$-DP, the worst-case subgroup vulnerability of any $\zval$, as well as magnitude of vulnerability disparity between any subgroups $\zval$ and $\zval'$, are uniformly bounded for any adversary's features $\obs$:
\begin{equation}\label{eq:dp-bound}
    \vulnz[\obs] \leq \exp(\varepsilon) - 1,\quad \big| \deltavuln[\obs] \big| \leq \exp(\varepsilon) - 1
\end{equation}
\end{proposition}

\begin{proof}
    Observe that the following probability distributions are equivalent:
    \begin{equation}\label{eq:neighbour-reformulation}
        \begin{aligned}
            \Pr_{\substack{\inset' \sim \datagen^{n - 1}\\ \xval \sim (\datagen \mid \zval)}} [\knowledge(\clf[\inset' \cup \{\xval\}], \xval)] &\equiv
            \Pr_{\substack{\inset \sim \datagen^n\\ \xval \sim (\inset \mid \zval)}}[\knowledge(\clf, \xval)] \\
            \Pr_{\substack{\inset' \sim \datagen^{n - 1}\\ \xval \sim (\datagen \mid \zval) \\ \xval' \sim \datagen}} [\knowledge(\clf[\inset' \cup \{\xval'\}], \xval)] &\equiv
            \Pr_{\substack{\inset \sim \datagen^n\\ \xval \sim (\datagen \mid \zval)}}[\knowledge(\clf, \xval)]
        \end{aligned}
    \end{equation}
    Notice that datasets $\inset' \cup \{\xval\}$ and $\inset' \cup \{\xval'\}$ differ by the records of at most one individual. Therefore, for any fixed dataset $\inset'$, the post-processing property of differential privacy applies:
    \[
        \begin{aligned}
        & \Pr_{\substack{\xval \sim (\datagen \mid \zval)}} [\knowledge(\clf[\inset' \cup \{\xval\}], \xval)] \leq \\ \leq
        & \exp(\varepsilon) 
        \Pr_{\substack{\xval \sim (\datagen \mid \zval) \\ \xval' \sim \datagen}} [\knowledge(\clf[\inset' \cup \{\xval'\}], \xval)]
        \end{aligned}
    \]
    Taking expectation over $\inset'$ of both sides, we obtain:
    \[
        \begin{aligned}
        & \Pr_{\substack{\inset' \sim \datagen^{n-1} \\ \xval \sim (\datagen \mid \zval)}} [\knowledge(\clf[\inset' \cup \{\xval\}], \xval)] \leq \\ \leq
        & \exp(\varepsilon) 
        \Pr_{\substack{\inset' \sim \datagen^{n-1} \\ \xval \sim (\datagen \mid \zval) \\ \xval' \sim \datagen}} [\knowledge(\clf[\inset' \cup \{\xval'\}], \xval)]
        \end{aligned}
    \]
    By equivalence in \cref{eq:neighbour-reformulation}:
    \[
        \Pr[\obs \mid \ismember = 1, \z = \zval] \leq \exp(\varepsilon) \Pr[\obs \mid \ismember = 0, \z = \zval]
    \]
    To get the bound on subgroup vulnerability, recall that by \cref{thm:subgroup-vuln} it is upper bounded by the total variation. Thus, for any set of feature values $T$:
    \[
        \begin{aligned}
            \vulnz[\obs] & \leq \sup_{T \subseteq \obspace} |\Pr[\obs \in T \mid \ismember = 1, \z = \zval] \\
            & - \Pr[\obs \in T \mid \ismember = 0, \z = \zval]| \\
            & \leq \exp(\varepsilon) - 1 \\
        \end{aligned}
    \]
    Applying \cref{thm:dv-upper-bound}, we also get the bound on disparity.
\end{proof}

\subsubsection{Empirical Evaluation}
To study how DP affects disparate vulnerability we train DP models with different
privacy levels. As target models,
we use DP logistic regression with private empirical risk minimization~\cite{ChaudhuriMS11}, trained
using the \textit{diffprivlib}~\cite{diffprivlib} implementation. 
We use a min-max scaler, and provide a maximum row norm estimated on a separate sample from the data distribution.
We use privacy levels $\varepsilon = 0.1, 1, 2, 10$.

We see in \cref{fig:sample-size-dp} that, for all evaluated values of $\varepsilon$, DP training considerably reduces disparity compared to the non-private logistic regression, with statistical tests not detecting significant deviations from 0. 

On the downside, unlike training with fairness constraints, DP training results in a significant decrease in accuracy of the models: from $45$ p.p. to $5$ p.p. drop depending on the value of $\varepsilon$.

\subsection{Takeaways}
Fairness only bounds disparate vulnerability in certain scenarios. Even when the classifier's fairness property generalizes beyond the training set, the bound is restricted to the adversarial strategy covered by the chosen fairness notion. Covering one adversarial strategy, however, is a weak security guarantee: the model could be (disparately) vulnerable to other strategies. Moreover, it is known that different fairness constraints are at odds with each other \cite{FriedlerSV16}. Hence, a model protected by one fairness notion may be inherently insecure against adversaries exploiting non-protected features.

Differential privacy bounds disparate vulnerability. We show that DP provides an upper bound on the vulnerability of all individuals, subgroups, and therefore on disparate vulnerability too. On the flip side, because DP guarantees are often at odds with accuracy, in practical applications $\varepsilon$ is usually set high, allowing for a lot of variation within the upper bound of \cref{thm:dp-bound-disparity}. Practically, the particular approach to DP training that we evaluated has mitigated disparity even with a high privacy level $\varepsilon = 10$ that results in vacuous theoretical bounds, but at significant accuracy costs.
\section{Evaluation using Real-World Datasets}
\label{sec:realworld}
To investigate if we can detect disparate vulnerability in a realistic setting, we use the following two datasets as case studies:
\begin{itemize}
    \item \textbf{\adult dataset}~\cite{Kohavi96}.
    The dataset contains 48,842 examples from the 1994
    Census database\footnote{\href{https://archive.ics.uci.edu/ml/datasets/adult}{https://archive.ics.uci.edu/ml/datasets/adult}}. The task is to determine if a yearly salary is over/under \$50K.
    It contains attributes such as age, sex, education, race, native country, etc.
    After one-hot encoding, the dataset contains 91 features. 
    We use the race column as the subgroup attribute.
    \item \textbf{\texas dataset.} We create this dataset based on 2013 Texas Hospital Discharge data\footnote{\href{https://www.dshs.texas.gov/THCIC/Hospitals/Download.shtm}{https://www.dshs.texas.gov/THCIC/Hospitals/Download.shtm}}.
    As our evaluation setup is computationally expensive, to accommodate the same training algorithms as used in the synthetic data experiments, we randomly subsample 50,000 examples, and reduce the number of features for training. We use the following columns: type of admission, illness severity, mortality risk, principal diagnosis code (out of more than 6000 codes, we only keep the top 1000 and create one separate code for the rest), length of stay, and patient's demographic attributes: sex, race, ethnicity. After one-hot encoding, we have 1025 features. We use the race column as the subgroup attribute.
    As a task, analogously to the \adult dataset, we use prediction of whether the total amount of charges is greater than a threshold (e.g., for health-insurance risk-scoring). As the threshold we pick the median total charges on the subsampled dataset. 
\end{itemize}

\cref{tab:datasets} provides details about the subgroups.

\begin{table}[t]
	\centering
	\caption{Subgroup representation in the datasets.
	}
	\label{tab:datasets}
	\begin{tabular}{llr} 
		Dataset & $\zval$ & Size \\
		\midrule
		\adult & ``White'' (WH) & 38,903 \\
		& ``Black'' (BL) & 4,228 \\
		& ``Asian-Pac-Islander'' (AI) & 1,303 \\
		& ``Amer-Indian-Eskimo'' (AE) & 435 \\
		& ``Other'' (OT) & 353 \\
		& All & 48,842 \\
		\midrule
        \texas & 4 &  31,514 \\
        & 5 &  10,883 \\
        & 3 &   6,451 \\
        & 2 &   1,019 \\
        & 1 &    133 \\
        & All  & 50,000 \\
	\end{tabular}
\end{table}

\begin{table*}[ht!]
    \centering
    \caption{Summary of models performance and vulnerability on \adult and \texas. Columns: \emph{Disparity test:} $p$-value of the \textsc{anova} F-test that checks if any of the subgroups have differing subgroup vulnerabilities, \emph{Test acc.:} Test accuracy of models, \emph{Gen. gap:} Per-model difference between train accuracy and test accuracy, \emph{Vuln.:} Aggregate vulnerability $\vulnadv[(\adv)]$. Bold font indicates models that have statistically significant disparity $(p < 0.01)$.}
    \label{tab:realworld-summary}
    \begin{tabular}{l|r|rr|rr|rr}
        {\adult} &     Disparity test & \multicolumn{2}{|c|}{Test acc.} & \multicolumn{2}{|c|}{Gen. gap} & \multicolumn{2}{|c}{Vuln., \%} \\
        {} &    $p$ &      avg &    std &  avg &    std & avg & std \\
        Model                    &        &           &        &          &        &                     &                    \\
        \midrule
        Logistic Regression (LR) & 0.3230 &    0.8404 & 0.0018 &   0.0012 & 0.0034 &              0.0942 &             0.4093 \\
        8-Neuron NN              & \textbf{0.0000} &    0.8421 & 0.0018 &   0.0044 & 0.0033 &              0.4052 &             0.3927 \\
        32-Neuron NN             & \textbf{0.0000} &    0.8410 & 0.0019 &   0.0131 & 0.0033 &              1.1373 &             0.4178 \\
        DP LR, $\varepsilon=1$          & 0.8534 &    0.7797 & 0.0135 &   0.0006 & 0.0040 &              0.0830 &             0.3478 \\
        DP LR, $\varepsilon=2$          & 0.0500 &    0.8053 & 0.0076 &   0.0004 & 0.0036 &              0.0563 &             0.3360 \\
        DP LR, $\varepsilon=10$         & 0.0419 &    0.8321 & 0.0023 &   0.0011 & 0.0032 &              0.0888 &             0.4100 \\
        Fair LR (Dem. Parity)    & 0.8945 &    0.8267 & 0.0018 &   0.0011 & 0.0035 &              0.0980 &             0.3331 \\
        Fair LR (Equalized Odds) & 0.7089 &    0.7941 & 0.0095 &   0.0006 & 0.0038 &              0.0782 &             0.3521 \\
    \end{tabular}
    \\\vspace{.1cm}
    \begin{tabular}{l|r|rr|rr|rr}
        {\texas} &     Disparity test & \multicolumn{2}{|c|}{Test acc.} & \multicolumn{2}{|c|}{Gen. gap} & \multicolumn{2}{|c}{Vuln., \%} \\
        {} &    $p$ &      avg &    std &  avg &    std & avg & std \\
        Model                    &        &           &        &          &        &                     &                    \\
        \midrule
        Logistic Regression (LR) & 0.2666 &    0.7833 & 0.0021 &   0.0152 & 0.0036 &              1.3905 &             0.4374 \\
        8-Neuron NN              & 0.0112 &    0.8836 & 0.0068 &   0.0282 & 0.0055 &              2.2384 &             0.5916 \\
        32-Neuron NN             & \textbf{0.0000} &    0.8639 & 0.0060 &   0.0686 & 0.0060 &              6.6238 &             0.7212 \\
        DP LR, $\varepsilon=1$   & 0.6192 &    0.6175 & 0.0191 &   0.0002 & 0.0045 &              0.0540 &             0.4317 \\
        DP LR, $\varepsilon=2$   & 0.0522 &    0.6363 & 0.0136 &   0.0014 & 0.0040 &              0.2125 &             0.3916 \\
        DP LR, $\varepsilon=10$   & 0.9737 &    0.7114 & 0.0146 &   0.0038 & 0.0041 &              0.5224 &             0.3245 \\
        Fair LR (Dem. Parity)    & \textbf{0.0078} &    0.7609 & 0.0028 &   0.0143 & 0.0039 &              1.2393 &             0.3444 \\
        Fair LR (Equalized Odds) & 0.7174 &    0.7477 & 0.0180 &   0.0133 & 0.0038 &              1.4676 &             0.3983 \\
    \end{tabular}
\end{table*}

\paragraph{Target models} We consider as target models logistic regression and neural networks with 8 and 32 neurons in the hidden layer (\cref{sec:experiments-main}), logistic regression with fairness constraints (\cref{sec:fairness}), and differentially private logistic regression with $\varepsilon$ values 1, 2, and 10 (\cref{sec:dp}). All the models beat the random accuracy baseline on the tasks.

\paragraph{Estimation method} As opposed to our synthetic data setup in which datasets to train shadow models can be directly sampled from the data-generating distribution, when real data is involved we can only sample data from the available finite dataset. We split the dataset in two parts: one for training of the shadow models, and one for evaluation of vulnerability~\cite{ShokriSSS17}. As a result, the amount of available training data is greatly reduced, in particular, for minority subgroups that already have few representatives in the dataset. To avoid this problem, in this section we use the average-threshold attack for vulnerability estimation, which does not require training shadow models. Our evaluation in \cref{sec:bias-eval} showed that this attack is not null-model biased. 

\paragraph{Setup} To train each target model, we randomly subsample 50\% of the dataset to use for training ($\inset_i$), and hold out the remaining data ($\outset_i$). We train 200 models for each model family on different splits of the dataset. For our statistical tests (see \cref{sec:stats}), we use $\alpha = 0.01$ as significance level.

\paragraph{Results} We summarize the results in \cref{tab:realworld-summary}. As in our synthetic experiments, we observe evidence of disparity in neural networks. Importantly, the results show that low vulnerability in absolute terms does not imply absence of disparity. On \adult, the 8-neuron network shows relatively low $0.4\%$ vulnerability but statistically significant disparity $(p < 10^{-4})$. Interestingly, on \texas, we also see statistical evidence of disparate vulnerability for logistic regression with demographic-parity constraints, although its overall vulnerability of 1.46\% is comparable to standard logistic regression. 

For the models with F-test $p < 0.01$, we conduct follow-up post-hoc tests to see which particular pairs of subgroups have high disparity (we defer the detailed results of the post-hoc tests to \cref{sec:extras}). On \adult, consistently with our synthetic experiments, the smaller subgroups ``Asian-Pac-Islander'' (AI, 1,302 examples), and ``Other'' (OT, 353 examples), exhibit disparity between themselves and other more populous subgroups. On \texas, almost all subgroup pairs exhibit significant disparity for 32-neuron network. 

The results for the logistic regression with fairness constraints are unlike the synthetic experiments. As opposed to a minority subgroup, as in the previous results, disparity appears between the most populous subgroup ``4'' (31,514 examples) and subgroups ``2'', ``3'' and ``5''.
This disparity does not exist in the standard logistic regression. Thus, this result shows that fairness constraints can introduce disparity when the conditions of \cref{thm:fairness-bound} are not met.

\paragraph{Discussion}
We have used binary classification tasks for compatibility with the fairness definitions, but we expect disparity to be more pronounced in multi-class settings.
As detailed in \cref{sec:vuln-takeaways}, disparate vulnerability is bound to happen whenever a model does not faithfully learn the distributional properties of the data for some subgroups.
Prior research suggests it is likely to appear when the task has many features, or many classes in the case of classification~\cite{Salem0HBF019}.

We also only considered relatively small dataset sizes. Bigger datasets, on the one hand, enable better learning of the models thus decreasing vulnerability and disparate vulnerability, but on the other hand, they would enable the adversary to use shadow-model attacks that could provide better results than the average-threshold attack used in our experiments. 

We leave for future work the investigation of the effect of number of classes and dataset size on disparate vulnerability.

\section{Conclusions}
\label{sec:discussion}
We have provided the first formal analysis of the disparate vulnerability of population
subgroups to membership inference attacks. Our analysis provides new insights into why and when vulnerability to MIAs arises and why and when these attacks have disparate impact. 

\paragraph{Key takeaways}
The first key learning of our study is that fully preventing MIAs, and thus preventing disparate vulnerability can only be done in two ways. Either by significantly increasing the complexity of the learning problem to ensure distributional generalization; or using a differentially-private training algorithm with the associated hit on performance.

The second learning surfaces a more general problem: the consequences of the unreliability of privacy estimation for demographic groups with a minority representation in the data. We show that for small subgroups it is easy to incorrectly estimate their protection indirectly via aggregate privacy measures, or directly when not considering biases adequately.

\paragraph{Why disparate vulnerability is important} Disparate vulnerability has crucial legal and policy significance.
Companies moving data between organizations or across borders face frictions designed to protect fundamental rights established by the approximately 140 countries with largely conceptually and textually similar privacy regulation around the world~\cite{GreenleafC20}.
For example, moving data from Europe into a country with significant state surveillance apparatus, such as the United States, is difficult after the European Court of Justice's judgement in \emph{Schrems II}.
Other countries, such as several in South Asia, have established specific personal data localization laws~\cite{BasuHS19}. 
As a consequence, there is growing interest in attempting to replace a direct trade in personal data with various forms of trade in models trained on this data.

Yet vulnerability of models to MIAs or other attacks compromising confidentiality might in some situations qualify models themselves as personal data~\cite{VealeBE18}.
The accountability principle in European data protection law places the onus on data controllers to demonstrate that a model should not be classified this way, for example through privacy-estimation techniques.
Our study indicates there is a real risk of ``privacy-washing'', laundering a model with aggregate statistics that mask vulnerabilities of subgroups.
It is true that prior work has also indicated that aggregate analysis can hide MIA vulnerability to attacks focusing on structurally vulnerable records~\cite{LongBWBWTGC18}.
However, this appears easier to dismiss as an acceptable residual leakage risk compared to disparate risks concerning members of salient minority groups, as in a liberal democracy, a regulator is more accountable towards these than towards a socially arbitrary selection of persons.

\paragraph{Open challenges}
Our results also uncover a new challenge. 
It is difficult for auditors or regulators to practically inspect disparate vulnerability, because they might lack a sufficient number of examples relating to a minority group.
When the subgroup data is scarce, our methods could be underpowered to detect disparity; however, not using the statistical tests and unbiased estimation methods from~\cref{sec:measuring} risks flagging disparity always when subgroups differ, devaluing the meaning of the estimate.

This points to a need for theoretical results that can be used as foundation in practical regulatory contexts.
Theoretical results may be able to help regulators better ascertain the limits of any metrics presented to them, and the conditions under which a model is structurally likely to be vulnerable to different types of privacy attacks even without difficult-to-obtain empirical evidence.
The initial results provided in this paper can already significantly contribute to discussions around the classification of machine learning systems in relation to their risk of data leakage as business practices of using models to transport information continue to evolve.
\section*{Acknowledgements}
The authors would like to thank Maksym Andriushchenko and Simon Oya for the helpful feedback and discussions. We also thank Hongyan Chang and Reza Shokri for clarifying their use of risk estimates~\cite{ChangS21}.

This work was partially funded by the Swiss National Science Foundation with grant 200021-188824.

\bibliographystyle{plainnat}
\bibliography{main}

\begin{thebibliography}{44}
\providecommand{\natexlab}[1]{#1}
\providecommand{\url}[1]{\texttt{#1}}
\expandafter\ifx\csname urlstyle\endcsname\relax
  \providecommand{\doi}[1]{doi: #1}\else
  \providecommand{\doi}{doi: \begingroup \urlstyle{rm}\Url}\fi

\bibitem[Bagdasaryan et~al.(2019)Bagdasaryan, Poursaeed, and
  Shmatikov]{BagdasaryanPS19}
Eugene Bagdasaryan, Omid Poursaeed, and Vitaly Shmatikov.
\newblock Differential privacy has disparate impact on model accuracy.
\newblock In \emph{Annual Conference on Neural Information Processing Systems,
  {NeurIPS}}, 2019.

\bibitem[Barocas and Selbst(2016)]{barocas_big_2016}
Solon Barocas and Andrew~D Selbst.
\newblock Big data's disparate impact.
\newblock \emph{Calif. L. Rev.}, 2016.

\bibitem[Basu et~al.(2019)Basu, Hickok, and Singh~Chawala]{BasuHS19}
Arindrajit Basu, Elonnai Hickok, and Aditya Singh~Chawala.
\newblock The {Localisation} {Gambit}: {Unpacking} {Policy} {Measures} for
  {Sovereign} {Control} of {Data} in {India}.
\newblock \emph{Centre for Internet and Society, India}, 2019.

\bibitem[Berk et~al.(2018)Berk, Heidari, Jabbari, Kearns, and
  Roth]{BerkHHJSKR18}
Richard Berk, Hoda Heidari, Shahin Jabbari, Michael Kearns, and Aaron Roth.
\newblock Fairness in criminal justice risk assessments: The state of the art.
\newblock \emph{Sociological Methods \& Research}, 2018.

\bibitem[Bird et~al.(2020)Bird, Dud{\'i}k, Edgar, Horn, Lutz, Milan, Sameki,
  Wallach, and Walker]{fairlearn}
Sarah Bird, Miro Dud{\'i}k, Richard Edgar, Brandon Horn, Roman Lutz, Vanessa
  Milan, Mehrnoosh Sameki, Hanna Wallach, and Kathleen Walker.
\newblock Fairlearn: A toolkit for assessing and improving fairness in {AI}.
\newblock Technical Report MSR-TR-2020-32, Microsoft, May 2020.
\newblock URL
  \url{https://www.microsoft.com/en-us/research/publication/fairlearn-a-toolkit-for-assessing-and-improving-fairness-in-ai/}.

\bibitem[Chang and Shokri(2021)]{ChangS21}
Hongyan Chang and Reza Shokri.
\newblock On the privacy risks of algorithmic fairness.
\newblock \emph{IEEE European Symposium on Security and Privacy, EuroS\&P},
  2021.

\bibitem[Chatzikokolakis et~al.(2020)Chatzikokolakis, Cherubin, Palamidessi,
  and Troncoso]{ChatzikokolakisCPT20}
Konstantinos Chatzikokolakis, Giovanni Cherubin, Catuscia Palamidessi, and
  Carmela Troncoso.
\newblock The {Bayes} security measure.
\newblock \emph{arXiv preprint arXiv:2011.03396}, 2020.

\bibitem[Chaudhuri et~al.(2011)Chaudhuri, Monteleoni, and
  Sarwate]{ChaudhuriMS11}
Kamalika Chaudhuri, Claire Monteleoni, and Anand~D. Sarwate.
\newblock Differentially private empirical risk minimization.
\newblock \emph{J. Mach. Learn. Res.}, 2011.

\bibitem[Cherubin et~al.(2019)Cherubin, Chatzikokolakis, and
  Palamidessi]{CherubinCP19}
Giovanni Cherubin, Konstantinos Chatzikokolakis, and Catuscia Palamidessi.
\newblock {F-BLEAU}: Fast black-box leakage estimation.
\newblock In \emph{IEEE Symposium on Security and Privacy, S\&P}, 2019.

\bibitem[Chouldechova(2017)]{chouldechova_fair_2016}
Alexandra Chouldechova.
\newblock Fair prediction with disparate impact: A study of bias in recidivism
  prediction instruments.
\newblock \emph{Big data}, 2017.

\bibitem[Chouldechova and Roth(2018)]{ChouldechovaR18}
Alexandra Chouldechova and Aaron Roth.
\newblock The frontiers of fairness in machine learning.
\newblock \emph{arXiv preprint arXiv:1810.08810}, 2018.

\bibitem[Devroye et~al.(2013)Devroye, Gy{\"o}rfi, and Lugosi]{DevroyeGL13}
Luc Devroye, L{\'a}szl{\'o} Gy{\"o}rfi, and G{\'a}bor Lugosi.
\newblock \emph{A probabilistic theory of pattern recognition}, volume~31.
\newblock Springer Science \& Business Media, 2013.

\bibitem[Dwork(2011)]{Dwork11}
Cynthia Dwork.
\newblock \emph{Differential Privacy}.
\newblock Springer {US}, 2011.

\bibitem[Dwork et~al.(2012)Dwork, Hardt, Pitassi, Reingold, and
  Zemel]{DworkHPRZ12}
Cynthia Dwork, Moritz Hardt, Toniann Pitassi, Omer Reingold, and Richard~S.
  Zemel.
\newblock Fairness through awareness.
\newblock In \emph{Innovations in Theoretical Computer Science}, 2012.

\bibitem[Ekstrand et~al.(2018)Ekstrand, Joshaghani, and
  Mehrpouyan]{EkstrandJM18}
Michael~D. Ekstrand, Rezvan Joshaghani, and Hoda Mehrpouyan.
\newblock Privacy for all: Ensuring fair and equitable privacy protections.
\newblock In \emph{Conference on Fairness, Accountability and Transparency,
  {FAT}}, 2018.

\bibitem[Farokhi and Kaafar(2020)]{farokhi2020modelling}
Farhad Farokhi and Mohamed~Ali Kaafar.
\newblock Modelling and quantifying membership information leakage in machine
  learning.
\newblock \emph{arXiv preprint arXiv:2001.10648}, 2020.

\bibitem[Friedler et~al.(2016)Friedler, Scheidegger, and
  Venkatasubramanian]{FriedlerSV16}
Sorelle~A Friedler, Carlos Scheidegger, and Suresh Venkatasubramanian.
\newblock On the (im) possibility of fairness.
\newblock \emph{arXiv preprint arXiv:1609.07236}, 2016.

\bibitem[Greenleaf and Cottier(2020)]{GreenleafC20}
Graham Greenleaf and Bertil Cottier.
\newblock 2020 ends a decade of 62 new data privacy laws.
\newblock \emph{Privacy Laws \& Business International Report}, 2020.

\bibitem[Hardt et~al.(2016)Hardt, Price, and Srebro]{HardtPNS16}
Moritz Hardt, Eric Price, and Nati Srebro.
\newblock Equality of opportunity in supervised learning.
\newblock In \emph{NIPS}, 2016.

\bibitem[Holohan et~al.(2019)Holohan, Braghin, Mac~Aonghusa, and
  Levacher]{diffprivlib}
Naoise Holohan, Stefano Braghin, P{\'o}l Mac~Aonghusa, and Killian Levacher.
\newblock Diffprivlib: The {IBM} differential privacy library.
\newblock \emph{arXiv preprint arXiv:1907.02444}, 2019.

\bibitem[Humphries et~al.(2020)Humphries, Rafuse, Tulloch, Oya, Goldberg,
  Hengartner, and Kerschbaum]{HumphriesRTOGK20}
Thomas Humphries, Matthew Rafuse, Lindsey Tulloch, Simon Oya, Ian Goldberg, Urs
  Hengartner, and Florian Kerschbaum.
\newblock Differentially private learning does not bound membership inference.
\newblock \emph{arXiv preprint arXiv:2010.12112}, 2020.

\bibitem[Jayaraman et~al.(2021)Jayaraman, Wang, Evans, and
  Gu]{jayaraman2020revisiting}
Bargav Jayaraman, Lingxiao Wang, David Evans, and Quanquan Gu.
\newblock Revisiting membership inference under realistic assumptions.
\newblock \emph{Proceedings on Privacy Enhancing Technologies}, 2021.

\bibitem[Kearns et~al.(1994)Kearns, Mansour, Ron, Rubinfeld, Schapire, and
  Sellie]{KearnsMRRSS94}
Michael~J. Kearns, Yishay Mansour, Dana Ron, Ronitt Rubinfeld, Robert~E.
  Schapire, and Linda Sellie.
\newblock On the learnability of discrete distributions.
\newblock In \emph{{ACM} Symposium on Theory of Computing}, 1994.

\bibitem[Khandani et~al.(2010)Khandani, Kim, and Lo]{KhandaniKL10}
Amir~E Khandani, Adlar~J Kim, and Andrew~W Lo.
\newblock Consumer credit-risk models via machine-learning algorithms.
\newblock \emph{Journal of Banking \& Finance}, 2010.

\bibitem[Kohavi(1996)]{Kohavi96}
Ron Kohavi.
\newblock Scaling up the accuracy of naive-bayes classifiers: A decision-tree
  hybrid.
\newblock In \emph{International Conference on Knowledge Discovery and Data
  Mining, KDD}, 1996.

\bibitem[Leino and Fredrikson(2020)]{LeinoF20}
Klas Leino and Matt Fredrikson.
\newblock Stolen memories: Leveraging model memorization for calibrated
  white-box membership inference.
\newblock In Srdjan Capkun and Franziska Roesner, editors, \emph{{USENIX}
  Security Symposium}, 2020.

\bibitem[Li et~al.(2021)Li, Li, and Ribeiro]{LiLR21}
Jiacheng Li, Ninghui Li, and Bruno Ribeiro.
\newblock Membership inference attacks and defenses in classification models.
\newblock In \emph{{CODASPY}}, 2021.

\bibitem[Lipton et~al.(2018)Lipton, McAuley, and Chouldechova]{LiptonMC18}
Zachary~C. Lipton, Julian McAuley, and Alexandra Chouldechova.
\newblock Does mitigating {ML}'s impact disparity require treatment disparity?
\newblock In \emph{Annual Conference on Neural Information Processing
  Systems,{NeurIPS}}, 2018.

\bibitem[Long et~al.(2020)Long, Wang, Bu, Bindschaedler, Wang, Tang, Gunter,
  and Chen]{LongBWBWTGC18}
Yunhui Long, Lei Wang, Diyue Bu, Vincent Bindschaedler, Xiaofeng Wang, Haixu
  Tang, Carl~A Gunter, and Kai Chen.
\newblock A pragmatic approach to membership inferences on machine learning
  models.
\newblock In \emph{IEEE European Symposium on Security and Privacy, EuroS\&P},
  2020.

\bibitem[Lum and Isaac(2016)]{LumIsaac16}
Kristian Lum and William Isaac.
\newblock To predict and serve?
\newblock \emph{Significance}, 2016.

\bibitem[Nakkiran and Bansal(2020)]{NakkiranBansal20}
Preetum Nakkiran and Yamini Bansal.
\newblock Distributional generalization: A new kind of generalization.
\newblock \emph{arXiv preprint arXiv:2009.08092}, 2020.

\bibitem[Nasr et~al.(2018)Nasr, Shokri, and Houmansadr]{NasrSH19}
Milad Nasr, Reza Shokri, and Amir Houmansadr.
\newblock Comprehensive privacy analysis of deep learning: Stand-alone and
  federated learning under passive and active white-box inference attacks.
\newblock In \emph{{IEEE} Symposium on Security and Privacy, {S\&P}}, 2018.

\bibitem[Obermeyer and Emanuel(2016)]{ObermeyerE16}
Ziad Obermeyer and Ezekiel~J Emanuel.
\newblock Predicting the future—big data, machine learning, and clinical
  medicine.
\newblock \emph{The New England journal of medicine}, 2016.

\bibitem[Pedregosa et~al.(2011)Pedregosa, Varoquaux, Gramfort, Michel, Thirion,
  Grisel, Blondel, Prettenhofer, Weiss, Dubourg, Vanderplas, Passos,
  Cournapeau, Brucher, Perrot, and Duchesnay]{scikit-learn}
F.~Pedregosa, G.~Varoquaux, A.~Gramfort, V.~Michel, B.~Thirion, O.~Grisel,
  M.~Blondel, P.~Prettenhofer, R.~Weiss, V.~Dubourg, J.~Vanderplas, A.~Passos,
  D.~Cournapeau, M.~Brucher, M.~Perrot, and E.~Duchesnay.
\newblock {Scikit-learn: Machine Learning in Python }.
\newblock \emph{Journal of Machine Learning Research}, 2011.

\bibitem[Pujol et~al.(2020)Pujol, McKenna, Kuppam, Hay, Machanavajjhala, and
  Miklau]{PujolMKHMM20}
David Pujol, Ryan McKenna, Satya Kuppam, Michael Hay, Ashwin Machanavajjhala,
  and Gerome Miklau.
\newblock Fair decision making using privacy-protected data.
\newblock In \emph{Conference on Fairness, Accountability, and Transparency,
  {FAT*}}, 2020.

\bibitem[Sablayrolles et~al.(2019)Sablayrolles, Douze, Schmid, Ollivier, and
  J{\'{e}}gou]{SablayrollesDSO19}
Alexandre Sablayrolles, Matthijs Douze, Cordelia Schmid, Yann Ollivier, and
  Herv{\'{e}} J{\'{e}}gou.
\newblock White-box vs black-box: Bayes optimal strategies for membership
  inference.
\newblock In \emph{International Conference on Machine Learning, {ICML}}, 2019.

\bibitem[Salem et~al.(2019)Salem, Zhang, Humbert, Berrang, Fritz, and
  Backes]{Salem0HBF019}
Ahmed Salem, Yang Zhang, Mathias Humbert, Pascal Berrang, Mario Fritz, and
  Michael Backes.
\newblock {ML}-leaks: Model and data independent membership inference attacks
  and defenses on machine learning models.
\newblock In \emph{26th Annual Network and Distributed System Security
  Symposium, {NDSS}}, 2019.

\bibitem[Seltman(2012)]{Seltman12}
Howard~J Seltman.
\newblock Experimental design and analysis.
\newblock 2012.

\bibitem[Shokri et~al.(2017)Shokri, Stronati, Song, and Shmatikov]{ShokriSSS17}
Reza Shokri, Marco Stronati, Congzheng Song, and Vitaly Shmatikov.
\newblock Membership inference attacks against machine learning models.
\newblock In \emph{{IEEE} Symposium on Security and Privacy, {S\&P}}, 2017.

\bibitem[Shokri et~al.(2019)Shokri, Strobel, and Zick]{ShokriSZ19}
Reza Shokri, Martin Strobel, and Yair Zick.
\newblock On the privacy risks of model explanations.
\newblock \emph{arXiv preprint arXiv:1907.00164}, 2019.

\bibitem[Song and Mittal(2021)]{SongMittal21}
Liwei Song and Prateek Mittal.
\newblock Systematic evaluation of privacy risks of machine learning models.
\newblock In \emph{{USENIX} Security Symposium}, 2021.

\bibitem[Veale et~al.(2018)Veale, Binns, and Edwards]{VealeBE18}
Michael Veale, Reuben Binns, and Lilian Edwards.
\newblock Algorithms that remember: model inversion attacks and data protection
  law.
\newblock \emph{Philosophical Transactions of the Royal Society A:
  Mathematical, Physical and Engineering Sciences}, 2018.

\bibitem[Yeom et~al.(2018)Yeom, Giacomelli, Fredrikson, and Jha]{YeomGFJ18}
Samuel Yeom, Irene Giacomelli, Matt Fredrikson, and Somesh Jha.
\newblock Privacy risk in machine learning: Analyzing the connection to
  overfitting.
\newblock In \emph{{IEEE} Computer Security Foundations Symposium, {CSF}},
  2018.

\bibitem[Zhao and Gordon(2019)]{ZhaoG19}
Han Zhao and Geoffrey~J. Gordon.
\newblock Inherent tradeoffs in learning fair representations.
\newblock In \emph{Annual Conference on Neural Information Processing Systems,
  {NeurIPS}}, 2019.

\end{thebibliography}

\appendix
\section{Proofs}
\label{sec:proofs}

In this section we provide the omitted proofs.

\subsection{Regular vs. Subgroup-Aware Vulnerability}

\paragraph{\cref{prop:discriminating-vs-regular}}\textit{
    $\vuln[\obs, \z] \geq \vuln[\obs]$.
}

\begin{proof}[Proof of \cref{prop:discriminating-vs-regular}]
    Recall that the Bayes adversary uses a Bayes-optimal classifier that maximizes the success
    probability (i.e., vulnerability) among all the possible classifiers.  That is, for the regular
    and subgroup-aware adversaries, we have respectively:

	\begin{align*}
	\vuln[\obs] &= \max_{g: \obspace \mapsto \{0, 1\}} \Pr[g(\yhat) = \ismember] \\
	\vuln[\obs, \z] &= \max_{g: \obspace \times \zset \mapsto \{0, 1\}} \Pr[g(\yhat, \z) = \ismember] \,.
	\end{align*}

	Let $F = \{f \mid f = g \circ h, h(\obsval, \zval) = \obsval,
	g: \obspace \mapsto \{0, 1\}\}$;
	that is, $F$ is the set of functions $f : \obspace \times \zset \mapsto \{0, 1\}$
	that first reduce the tuple $(\obsval, \zval)$ to $\obsval$
	and then apply a function $g$ to the remaining input.
	Clearly, $F \subset \{g \mid g:\obspace \times \zset \mapsto \{0, 1\}\}$.

	Then, to prove this proposition it suffices to observe that
	the regular adversary is equivalent to a subgroup-aware one
	restricted to the set of functions $F$.

	\begin{align*}
	\vuln[\obs, \z] &= \max_{g: \obspace \times \zset \mapsto \{0, 1\}} \Pr[g(\obsval, \z) = \ismember]\\
	&\geq \max_{f \in F} \Pr[f(\obsval, \z) = \ismember]\\
	&= \max_{g: \obspace \mapsto \{0, 1\}} \Pr[g(\obsval) = \ismember]\\
	&= \vuln[\obs] \,.
	\end{align*}
\end{proof}

\subsection{Subgroup Vulnerability}

\begin{table*}[h]
    \centering
    \caption{Results of post-hoc tests on \adult models. Columns: \emph{$\zval$ and $\zval'$:} identifiers of subgroups, \emph{$t$:} value of the t statistic, \emph{$p$:} uncorrected p-value, \emph{$p$-corr.:} p-value after the correction for multiple comparisons.}
    \label{tab:adult-posthocs}
    \begin{tabular}{lllrrrl}
    {NN-8} &              $\zval$ &              $\zval'$ &    $t$ &   $p$ &  $p$-corr. \\
    \midrule
    0 &     AE &     AI & -4.4298 & 0.0000 &     \textbf{0.0001} \\
    1 &     AE &     BL &  0.5143 & 0.6076 &     0.6751 \\
    2 &     AE &     OT & -1.7468 & 0.0822 &     0.1174 \\
    3 &     AE &     WH &  0.0498 & 0.9604 &     0.9604 \\
    4 &     AI &     BL &  8.8677 & 0.0000 &     \textbf{0.0000} \\
    5 &     AI &     OT &  1.8976 & 0.0592 &     0.0987 \\
    6 &     AI &     WH &  8.9236 & 0.0000 &     \textbf{0.0000} \\
    7 &     BL &     OT & -2.6402 & 0.0089 &     0.0224 \\
    8 &     BL &     WH & -1.3443 & 0.1804 &     0.2255 \\
    9 &     OT &     WH &  2.3290 & 0.0209 &     0.0417 \\
    \end{tabular}~
    \begin{tabular}{lllrrrl}
    {NN-32} &              $\zval'$ &              $\zval'$ &     $t$ &   $p$ &  $p$-corr. \\
    \midrule
    0 &     AE &     AI & -11.3216 & 0.0000 &     \textbf{0.0000} \\
    1 &     AE &     BL &   0.9595 & 0.3385 &     0.3761 \\
    2 &     AE &     OT &  -4.1972 & 0.0000 &     \textbf{0.0001} \\
    3 &     AE &     WH &   0.5655 & 0.5724 &     0.5724 \\
    4 &     AI &     BL &  24.1213 & 0.0000 &     \textbf{0.0000} \\
    5 &     AI &     OT &   6.1285 & 0.0000 &     \textbf{0.0000} \\
    6 &     AI &     WH &  25.4526 & 0.0000 &     \textbf{0.0000} \\
    7 &     BL &     OT &  -6.4301 & 0.0000 &     \textbf{0.0000} \\
    8 &     BL &     WH &  -1.2845 & 0.2005 &     0.2506 \\
    9 &     OT &     WH &   6.1996 & 0.0000 &     \textbf{0.0000} \\
    \end{tabular}
\end{table*}

To prove \cref{thm:subgroup-vuln}, we use the following statement:
\begin{proposition}\label{thm:half-tv}
    For any two discrete probability measures $\mu$ and $\mu'$ the following holds:
    \[
        \sum_{\xval : \mu(\xval) > \mu'(\xval)} \left[ \mu(\xval) - \mu'(\xval) \right] = \frac{1}{2}||\mu - \mu'||_1.
    \]
\end{proposition}
\begin{proof}
First, observe:
\[
    \begin{aligned}
        \frac{1}{2}||\mu - \mu'||_1 &= \frac{1}{2} \sum_{\xval} |\mu(x) - \mu'(x)| \\
        & = \frac{1}{2} \sum_{\mu(x) > \mu'(x)} (\mu(x) - \mu'(x)) \\
            &- \frac{1}{2} \sum_{\mu(x) \leq \mu'(x)} (\mu(x) - \mu'(x)).
    \end{aligned}
\]
Rearranging and grouping the terms, we get:
\[
    \begin{aligned}
    & = \frac{1}{2}\Big(\sum_{\mu(x) > \mu'(x)} \mu(x) - \sum_{\mu(x) \leq \mu(x)} \mu'(x) \\
                         &\quad- \sum_{\mu(x) > \mu'(x)} \mu'(x) + \sum_{\mu(x) \leq \mu'(x)}
                         \mu'(x) \Big) \\
        & = \sum_{\xval : \mu(\xval) > \mu'(\xval)} \left[ \mu(\xval) - \mu'(\xval) \right]
    \end{aligned}
\]
\end{proof}

\begin{proof}[Proof of \cref{thm:subgroup-vuln}]
    We provide a proof for the case of discrete features $\obs$. The proof is analogous in the case of absolutely continuous $\obs$.
    Note that for discrete measures $\mu$ and $\mu'$, $\TV(\mu, \mu') = \frac{1}{2} || \mu - \mu'||_1$. 
    
    For convenience, let us define feature gaps as follows:
    \[
        \begin{aligned}
            \gap(\obsval) &\define \mu_{1}(\obsval) - \mu_{0}(\obsval) \\
            \gap_\zval(\obsval) &\define \mu_{1,\zval}(\obsval) - \mu_{0,\zval}(\obsval)
        \end{aligned}
    \]
    Adversary's success for a subgroup has the following form that is useful for our proof:
    \begin{equation}\label{eq:subgroup-vuln-aux}
        \begin{aligned}
            & 2 \Pr[\attbayes(\obs) = \ismember \mid \z = \zval] - 1 = \\
            & = \Pr[\attbayes(\obs) = 1 \mid \ismember=1, \z = \zval] \\
            & - \Pr[\attbayes(\obs) = 1 \mid \ismember=0, \z = \zval] \\
            & = \sum_{\obsval: \attbayes(\obsval) = 1} \mu_{1,\zval}(\obsval)
              + \sum_{\obsval: \attbayes(\obsval) = 1} \mu_{0,\zval}(\obsval) \\
            & = \sum_{\obsval: \mu_{1}(\obsval) > \mu_{0}(\obsval)} \big[ \mu_{1,\zval}(\obsval)
              - \mu_{0,\zval}(\obsval)\big] \\
            & = \sum_{\obsval: \gap(\obsval) > 0}
                \gap_\zval(\obsval)
        \end{aligned}
    \end{equation}
    First, suppose that $\z \notin \obs$. Consider the following set:
    \[
        \begin{aligned}
            C &\define \{\obsval \mid \gap(\obsval) > 0\} =  \{\obsval \mid \mu_1(\obsval) > \mu_0(\obsval)\}
        \end{aligned}
    \]
    For a given $\zval$, the set $C$ is a union of two other disjoint sets $A$ and $B$; $C = A \cup B$:
    \[
        \begin{aligned}
            A &= \{ \obsval \mid \mu_{1,\zval}(\obsval) \leq \mu_{0,\zval}(\obsval) \land
            \mu_{1}(\obsval) > \mu_{0}(\obsval) \} \\
            B &= \{ \obsval \mid \mu_{1,\zval}(\obsval) > \mu_{0,\zval}(\obsval) \land
            \mu_{1}(\obsval) > \mu_{0}(\obsval) \}
        \end{aligned}
    \]
    Thus, the sum in Eq.~\ref{eq:subgroup-vuln-aux} can be decomposed into $\sum_A \gap_\zval(\obsval) +
    \sum_B \gap_\zval(\obsval)$, where
    \[
        \begin{aligned}
            & \sum_A \gap_\zval(\obsval)
            = \sum_{\gap_\zval(\obsval) \leq 0 \land \cdots} \gap_\zval(\obsval)
            \leq 0 \\
            0 \leq & \sum_B \gap_\zval(\obsval)
            \leq \sum_{\gap_\zval(\obsval) > 0} \gap_{\zval}(\obsval) = \frac{1}{2}||\mu_{1,\zval} - \mu_{0,\zval}||_1
        \end{aligned}
    \]
    The last equality is by \cref{thm:half-tv}. Applying this bound to
    \cref{eq:subgroup-vuln-aux} we obtain the sought \cref{eq:subgroup-vuln}.

    \smallskip

    Second, suppose that $\z \in \obs$. Let $\obsval = (\cdots, \zval')$. If $\zval' \neq \zval$,
    then $\gap_\zval(\obsval) = 0$, and so we only need to consider the case $\zval' = \zval$. In this case:
    \[
        \begin{aligned}
            \id[\gap(\obsval) > 0]
            &= \id[\mu_1(\obsval) > \mu_0(\obsval)] \\
            &= \id\big[\mu_{1,\zval}(\obsval) \cdot \Pr[\zval] > \mu_{0,\zval}(\obsval) \cdot \Pr[\zval] \big] \\
            &= \id[\gap_\zval(\obsval) > 0].
        \end{aligned}
    \]
    After plugging this into \cref{eq:subgroup-vuln-aux}, we obtain the equality in
    \cref{eq:subgroup-vuln-disc} by \cref{thm:half-tv}.
\end{proof}

\subsection{A Note on Equalized Odds vs. Demographic Parity}
Let us define equalized odds (EO). With probabilities over the data distribution, a classifier satisfies EO if:
\[
    \Pr[\yhat \mid \y, \z = \zval] = \Pr[\yhat \mid \y, \z = \zval']
\]
In these terms, demographic parity is defined as the following requirement for a classifier:
\[
    \Pr[\yhat \mid \z = \z] = \Pr[\yhat \mid \z = \z']
\]
In general, these two notions are not equivalent. In our synthetic data setup (\cref{sec:bias-eval}), however, it holds that (a) the distributions of classes are the same across subgroups: $\Pr[\y \mid \z = \z] = \Pr[\y \mid \z = \z']$, and (b) the two classes are balanced: $\Pr[\y = 1] = \Pr[\y = 0] = \nicefrac{1}{2}$. It is easy to see that in this case, EO implies demographic parity.

\section{Additional Tables}
\label{sec:extras}
The rest of the appendix contains additional tables.

\begin{table*}[h]
    \centering
    \caption{Results of post-hoc tests on \texas models. See \cref{tab:adult-posthocs} caption for details.}
    \label{tab:texas-posthocs}
    \begin{tabular}{lllrrrl}
    {NN-32} &              $\zval$ &              $\zval'$ &    $t$ &   $p$ &  $p$-corr. \\
    \midrule
    0 &      1 &      2 & -3.4973 & 0.0006 &     \textbf{0.0007} \\
    1 &      1 &      3 &  0.2056 & 0.8374 &     0.8374 \\
    2 &      1 &      4 &  4.2820 & 0.0000 &     \textbf{0.0000} \\
    3 &      1 &      5 &  3.0576 & 0.0025 &     \textbf{0.0028} \\
    4 &      2 &      3 & 10.0174 & 0.0000 &     \textbf{0.0000} \\
    5 &      2 &      4 & 21.2727 & 0.0000 &     \textbf{0.0000} \\
    6 &      2 &      5 & 17.4069 & 0.0000 &     \textbf{0.0000} \\
    7 &      3 &      4 & 21.8804 & 0.0000 &     \textbf{0.0000} \\
    8 &      3 &      5 & 13.2434 & 0.0000 &     \textbf{0.0000} \\
    9 &      4 &      5 & -8.1600 & 0.0000 &     \textbf{0.0000} \\
    \end{tabular}~
    \begin{tabular}{lllrrrl}
    {LR (Dem. Parity)} &              $\zval'$ &              $\zval'$ &     $t$ &   $p$ &  $p$-corr. \\
    \midrule
    0 &      1 &      2 & -1.2485 & 0.2133 &     0.3326 \\
    1 &      1 &      3 & -1.1910 & 0.2351 &     0.3326 \\
    2 &      1 &      4 & -2.4808 & 0.0139 &     0.0348 \\
    3 &      1 &      5 & -0.9385 & 0.3491 &     0.3879 \\
    4 &      2 &      3 &  0.3151 & 0.7531 &     0.7531 \\
    5 &      2 &      4 & -3.4931 & 0.0006 &     \textbf{0.0020} \\
    6 &      2 &      5 &  1.1152 & 0.2661 &     0.3326 \\
    7 &      3 &      4 & -8.8594 & 0.0000 &     \textbf{0.0000} \\
    8 &      3 &      5 &  1.6787 & 0.0948 &     0.1896 \\
    9 &      4 &      5 & 12.8701 & 0.0000 &     \textbf{0.0000} \\
    \end{tabular}
\end{table*}

\begin{table*}[h!]
    \centering
    \caption{Results on \adult, disaggregated by subgroups, for models with disparity F-test $p < 0.01$.}
    \label{tab:adult-disaggregated}
    \begin{tabular}{llrrrrrr}
                          &   & \multicolumn{2}{l}{Test acc.} & \multicolumn{2}{l}{Gen. gap} & \multicolumn{2}{l}{Subgroup vuln.} \\
                          &   &      avg &    std &     avg &    std & avg & std \\
    Model & $\zval$ &           &        &          &        &                     &                    \\
    \midrule
    32-Neuron NN & Amer-Indian-Eskimo &    0.9028 & 0.0139 &   0.0115 & 0.0253 &              1.1701 &             4.8259 \\
                             & Asian-Pac-Islander &    0.8165 & 0.0119 &   0.0693 & 0.0195 &              5.7713 &             2.6300 \\
                             & Black &    0.9043 & 0.0049 &   0.0138 & 0.0086 &              0.8200 &             1.6261 \\
                             & Other &    0.8881 & 0.0179 &   0.0492 & 0.0295 &              3.2550 &             5.1807 \\
                             & White &    0.8338 & 0.0021 &   0.0109 & 0.0035 &              0.9773 &             0.4496 \\
    8-Neuron NN & Amer-Indian-Eskimo &    0.9042 & 0.0151 &   0.0041 & 0.0281 &              0.3701 &             4.7177 \\
                             & Asian-Pac-Islander &    0.8264 & 0.0119 &   0.0223 & 0.0214 &              2.1320 &             2.7965 \\
                             & Black &    0.9066 & 0.0047 &   0.0035 & 0.0093 &              0.1878 &             1.6152 \\
                             & Other &    0.8913 & 0.0165 &   0.0149 & 0.0309 &              1.2805 &             5.6344 \\
                             & White &    0.8345 & 0.0020 &   0.0039 & 0.0036 &              0.3535 &             0.4314 \\
    \end{tabular}
\end{table*}

\begin{table*}[h]
    \caption{Results on \texas, disaggregated by subgroups, for models with disparity F-test $p < 0.01$.}
    \label{tab:texas-disaggregated}
    \centering
    \begin{tabular}{llrrrrrr}
                          &   & \multicolumn{2}{l}{Test acc.} & \multicolumn{2}{l}{Gen. gap} & \multicolumn{2}{l}{Subgroup vuln.} \\
                          &   &      avg &    std &     avg &    std & avg & std \\
    Model & $\zval$ &           &        &          &        &                     &                    \\
    \midrule
    32-Neuron NN & 1 &    0.8699 & 0.0380 &   0.0791 & 0.0451 &              8.5188 &             8.2829 \\
                          & 2 &    0.8644 & 0.0153 &   0.1013 & 0.0180 &             10.7429 &             3.0129 \\
                          & 3 &    0.8498 & 0.0085 &   0.0855 & 0.0106 &              8.3947 &             1.6121 \\
                          & 4 &    0.8644 & 0.0066 &   0.0637 & 0.0063 &              6.0331 &             0.8261 \\
                          & 5 &    0.8708 & 0.0063 &   0.0697 & 0.0074 &              6.7288 &             1.0840 \\
    Fair LR (Dem. Parity) & 1 &    0.6932 & 0.0562 &  -0.0010 & 0.0839 &              0.0075 &             8.9200 \\
                          & 2 &    0.6934 & 0.0203 &   0.0095 & 0.0295 &              0.8381 &             2.9201 \\
                          & 3 &    0.7323 & 0.0084 &   0.0143 & 0.0099 &              0.7667 &             1.1361 \\
                          & 4 &    0.7771 & 0.0027 &   0.0155 & 0.0048 &              1.5751 &             0.4952 \\
                          & 5 &    0.7384 & 0.0068 &   0.0106 & 0.0088 &              0.5997 &             0.8448 \\
    \end{tabular}
\end{table*}

\end{document}